%% file: acl_latex.tex
\documentclass[11pt]{article}

\usepackage[final]{acl}

\usepackage{times}
\usepackage{latexsym}

\usepackage[T1]{fontenc}

\usepackage[utf8]{inputenc}

\usepackage{microtype}

\usepackage{inconsolata}

\usepackage{graphicx}
\usepackage{float}

\usepackage{tcolorbox}
\usepackage{listings}
\tcbuselibrary{breakable,listings}
\usepackage{xcolor}
\definecolor{famForecast}{HTML}{F0F9FF}
\definecolor{famData}{HTML}{ECFDF5}
\definecolor{famAnomaly}{HTML}{FFFBEB}
\definecolor{famDecision}{HTML}{F5F3FF}
\definecolor{famRoot}{HTML}{FFFFFF}
\usepackage{amsmath}
\usepackage{amsfonts}
\usepackage{amssymb}
\usepackage{booktabs}
\usepackage{array}
\usepackage[edges]{forest}
\usepackage{tikz}
\usetikzlibrary{arrows.meta,calc}

\forestset{
  tsTax/.style={
    for tree={
      grow=east,
      parent anchor=east,
      child anchor=west,
      reversed,
      align=left,
      text ragged,
      draw=blue!65!black,
      rounded corners=2pt,
      thick,
      inner xsep=4pt,
      inner ysep=3pt,
      font=\scriptsize,
      edge={gray!70, very thick},
      edge path={
        \noexpand\path [\forestoption{edge}]
          (!u.east) -- ++(3pt,0pt) |- (.west) -- (.west);
      },
      l sep=6pt,
      s sep=4pt,
    },
  },
  tsRoot/.style={
    draw=blue!75!black,
    very thick,
    rounded corners=3pt,
    fill=famRoot,
    font=\scriptsize\bfseries,
    align=center,
    text width=1.7cm,
  },
  tsMid/.style={
    font=\scriptsize\bfseries,
    text width=2.2cm,
  },
  tsLeafNarrow/.style={
    font=\scriptsize,
    text width=1.5cm,
  },
  tsLeaf/.style={
    font=\scriptsize,
    text width=8.9cm,
  },
  tsFamForecast/.style={for tree={fill=famForecast}},
  tsFamData/.style={for tree={fill=famData}},
  tsFamAnomaly/.style={for tree={fill=famAnomaly}},
  tsFamDecision/.style={for tree={fill=famDecision}},
}

\title{LLM Agents for Time-Series: A Survey}

\author{%
  \textbf{Yilong Chen$^{1}$ \quad Xiao Qin$^{1}$ \quad Chenghao Liu$^{2}$\footnotemark} \\
  \textbf{Liang Wu$^{3}$ \quad Noelle I.\ Samia$^{1}$ \quad Kaize Ding$^{1\dagger}$} \\
  $^{1}$Northwestern University \quad
  $^{2}$Datadog AI Research \quad
  $^{3}$Nokia \\
  {\small $^{\dagger}$Corresponding author}
}

\begin{document}
\begingroup
\renewcommand{\thefootnote}{*}
\maketitle
\footnotetext{This work was completed prior to joining Datadog.}
\endgroup
\setcounter{footnote}{0}

\input{0_abstract}


\input{1_introduction}
\input{2_background}
\input{3_architectures}
\input{4_0_taxonomy}
\input{4_1_taxonomy}
\input{4_2_taxonomy}
\input{4_3_taxonomy}
\input{4_4_taxonomy}
\input{5_resources}
\input{6_future}

\section*{Limitations}

While this survey provides a comprehensive overview of LLM-based agentic systems for time-series tasks, it has several limitations:

\noindent\textbf{Scope and coverage.} Due to the rapid pace of advancements in LLM-based agents, some recent developments and emerging directions may not be fully captured in this survey.

\noindent\textbf{Lack of quantitative comparison.} The broad range of time-series tasks and heterogeneous evaluation settings make it difficult to establish a unified and fair empirical comparison across all systems.

\noindent\textbf{Design guidance.} The design insights are derived from recurring patterns observed in the literature rather than controlled experimental validation, and thus may not generalize to all practical settings.

\noindent Despite these limitations, we hope this survey provides a useful and structured reference for understanding and designing LLM-based time-series agents.

\bibliography{anthology,custom}

\clearpage
\appendix

\newif\ifshowappendixfloats
\showappendixfloatsfalse

\input{appendix_01_survey}
\input{appendix_03_prompts}
\input{appendix_04_patterns}
\input{appendix_07_failure_modes}
\input{appendix_05_resources}
\input{appendix_06_evaluation}

\clearpage
\showappendixfloatstrue
\makeatletter
\setlength{\@fptop}{0pt}
\setlength{\@fpbot}{0pt plus 1fil}
\setlength{\@dblfptop}{0pt}
\setlength{\@dblfpbot}{0pt plus 1fil}
\makeatother

\input{appendix_02_methods}

\input{appendix_04_patterns}
\input{appendix_05_resources}
\input{appendix_06_evaluation}
\end{document}

%% file: 0_abstract.tex
\begin{abstract}

LLM-based agents are increasingly being developed for time-series problems, but their design choices vary substantially across task settings.
This survey adopts a problem-driven taxonomy that organizes these systems by the time-series problems they address rather than by isolated technical components.
We group existing systems into four categories: forecasting and reasoning, augmentation and synthesis, anomaly detection and diagnosis, and decision support.
Within each category, we examine how task requirements shape agent architecture, tool use, and memory design.
We further summarize representative datasets and environments, and compare reported model performance under shared or closely related settings.
Overall, this survey offers a task-oriented guide to designing LLM-based agents for time-series problems and identifies open gaps for future work.

\end{abstract}

%% file: 1_introduction.tex
\section{Introduction}
\label{sec:introduction}

Time-series analysis~\cite{hamilton2020time} plays an important role in many real-world domains, including finance~\cite{tsay2005analysis}, transportation~\cite{xu2016big}, and climate science~\cite{kim2025climateagent}. Representative tasks include forecasting~\cite{chatfield2000time,de200625}, data augmentation~\cite{wen2020time}, anomaly detection~\cite{blazquez2021review,schmidl2022anomaly}, and decision support. These tasks have long been addressed with statistical and classical machine learning models such as ARIMA~\cite{shumway2017arima}, bootstrapping~\cite{efron1992bootstrap}, and LOF~\cite{breunig2000lof}, and more recently with deep models such as RNNs~\cite{medsker2001recurrent} and Transformers~\cite{vaswani2017attention}.

\begin{figure}[t]
    \centering
    \includegraphics[width=\columnwidth]{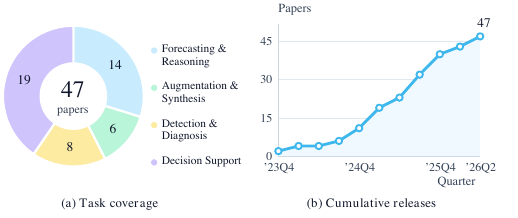}
    \caption{Task coverage and release timeline of 47 surveyed LLM-agent papers for time-series tasks.}
    \label{fig:teaser_overview}
\end{figure}

However, applying these methods often relies heavily on expert knowledge to design analysis pipelines and interpret results. Recent advances in large language models (LLMs)~\cite{zhao2023survey} offer an alternative by leveraging general-purpose language reasoning to assist time-series analysis, for example as modules for encoding and explaining temporal patterns. Moving beyond prompt-only use, LLMs can be deployed as \emph{agents}~\cite{wang2024survey} that plan actions, call tools, and maintain memory across multiple steps. Such agentic systems are particularly suitable for time-series applications because many time-series tasks require updates from new observations, decisions under evolving conditions, and tool-augmented integration of heterogeneous evidence rather than a fixed pipeline~\cite{cheng2026atsf,tao2026castr1,zhangprompts}. Motivated by this shift, this survey reviews recent progress on LLM-based agentic systems for time-series tasks.

Recent time-series leaderboards further suggest that agentic systems are becoming competitive with state-of-the-art methods by harnessing strong time-series models rather than replacing them~\cite{aksu2024gift}. By coordinating foundation models, retrieval, validation, and domain tools within task-specific workflows, these systems shift the practical question from which model to use toward how agent architectures, tools, and memory should be designed for each task. This motivates our problem-driven taxonomy of time-series agentic systems.

\noindent\textbf{Positioning.} Existing LLM-for-time-series surveys~\cite{chang2025survey,zhangprompts} typically organize methods by individual LLM capabilities (e.g., planning, reasoning, memory, and tool use), rather than by how these capabilities are integrated for concrete time-series tasks. Conversely, domain-general LLM-agent surveys~\cite{wang2024survey,guo2024large,li2024survey,ferrag2025llm} seldom address time-series-specific challenges such as streaming inputs, temporal dependence, distribution shift, and action--data coupling. We therefore adopt a \emph{problem-driven taxonomy} that groups methods by target task and analyzes recurring design patterns in architecture, tool use, and memory. A detailed comparison with related surveys is provided in Appendix~\ref{sec:appendix_survey_comparison}.

The remainder of this survey introduces the necessary background and core design dimensions in Sections~2--3, presents the problem-driven taxonomy in Section~4, and reviews practical resources and future research directions in Sections~5--6.


%% file: 2_background.tex
\section{Background and Foundations}

\noindent\textbf{Time-Series Data and Representations.} A time series may arise from a continuous-time process, but practical systems usually operate on observed or tokenized indices. Let
$\{\tau_t\}_{t=1}^T$ satisfy $\tau_1 < \cdots < \tau_T$, and denote
$$
\mathbf{X}_{\tau_1:\tau_T} = (\mathbf{x}_{\tau_1}, \dots, \mathbf{x}_{\tau_T}), 
\quad \mathbf{x}_{\tau_t} \in \mathbb{R}^d ,
$$
where $d$ covers observed channels, covariates, or spatially distributed sensors.
For irregular, noisy, or partially observed data, systems typically construct
$\mathbf{z}_{\tau_t} = \phi(\mathbf{X}_{\tau_1:\tau_t})$
to summarize historical context.

\noindent\textbf{Time-Series Modeling.} Time-series modeling transforms temporal observations into forecasts, anomaly scores, learned representations, or explanatory summaries. Statistical models (e.g., ARIMA, ARCH/GARCH, HMMs) encode assumptions about dependence, stationarity, latent regimes, and uncertainty~\cite{shumway2017arima,engle1982arch,bollerslev1986garch,rabiner1989hmm}. Deep models (e.g., LSTM, TCN, DeepAR) learn temporal patterns~\cite{hochreiter1997lstm,bai2018tcn,salinas2020deepar}, and LLM-based models (e.g., Time-LLM, LSTPrompt) represent sequences through tokenized or multimodal inputs for language-based reasoning or explanation~\cite{jin2023timellm,liu2024lstprompt}. As standalone components, however, these models typically do not choose workflows, call tools, or revise memory and hypotheses based on intermediate feedback.

\noindent\textbf{From Inference to Agentic Systems.} Static analysis pipelines are often inadequate for interactive time-series tasks that require evidence gathering, feedback integration, or adaptive action selection. We view LLM agents as observe--act--update systems that combine reasoning, planning, tool use, and memory~\cite{wang2024survey, zhangprompts}, and use this distinction to separate agentic systems from prompt-only LLM applications.

%% file: 3_architectures.tex
\section{Fundamental Design Dimensions}
\label{sec:agent_architectures}

Before the task taxonomy, we summarize three design dimensions of time-series agentic systems: architecture, tools, and memory.

\subsection{Architecture}

We distinguish single-agent and multi-agent architectures; for multi-agent systems, we further describe their architectural patterns as cooperative, competitive, or mixed~\cite{zhu2024surveycommunication}.

\begin{figure}[H]
  \centering
  \includegraphics[width=\linewidth]{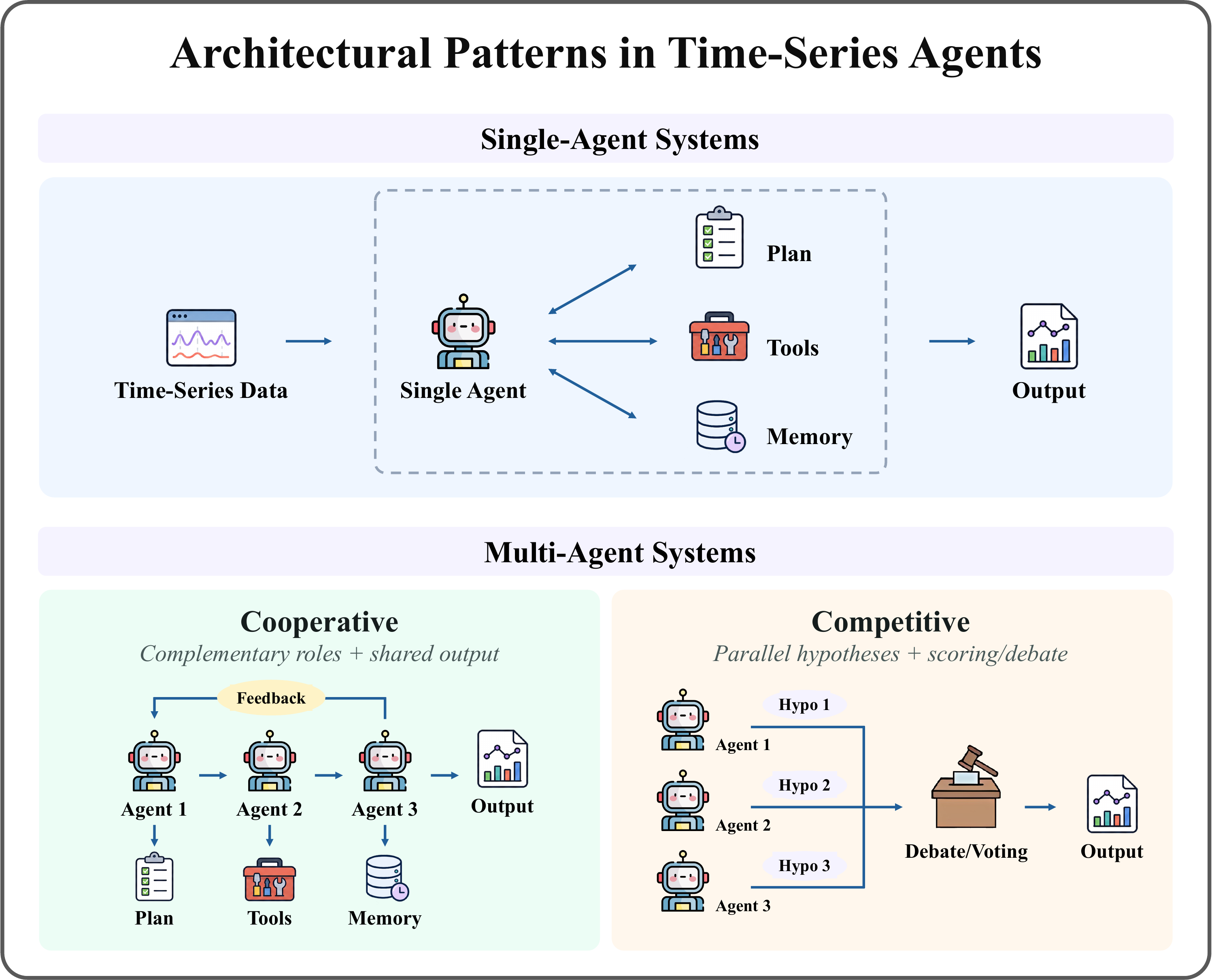}
  \caption{An illustration of the main architectural patterns for time-series agent systems.}
  \label{fig:architecture_overview}
\end{figure}

\noindent\textbf{Single-Agent Systems.}
A single-agent system uses one agent to plan reasoning steps, call tools, maintain memory, and produce outputs or actions from current observations~\cite{yao2023reactsynergizingreasoningacting}.

\noindent\textbf{Cooperative Multi-Agent Systems.}
Cooperative systems assign agents complementary roles or subtasks that are not directly interchangeable, such as sequential stages or planner--executor structures, so agents coordinate toward a shared output~\cite{li2024survey,wooldridge2009multiagent}.

\noindent\textbf{Competitive Multi-Agent Systems.}
Competitive systems instantiate agents or agent-generated candidates as alternatives: they produce competing hypotheses, forecasts, explanations, rules, or actions, which are then resolved by scoring, ranking, voting, debate, or an explicit judge~\cite{zhu2024surveycommunication}.

\noindent\textbf{Mixed Multi-Agent Systems.}
Mixed systems combine cooperative role decomposition with competitive comparison, debate, or selection within the same workflow~\cite{zhu2024surveycommunication}.

\subsection{Tools}
\label{sec:agent_tools}

Tools are callable interfaces to external programs invoked by the LLM agent~\cite{wang2024tools}. In time-series agents, they enable access to temporal data, specialized computation, and verifiable feedback that cannot be obtained reliably from parametric knowledge alone. Common tool families include \emph{(i) Database APIs}, which provide structured access to stored data; \emph{(ii) Search \& Retrieval APIs}, which return relevant external information or historical records; \emph{(iii) Data Processing Tools}, which transform raw inputs into more useful representations, such as through alignment or dynamic time warping (DTW); \emph{(iv) Statistical \& ML Models}, which produce predictions, scores, or learned representations; and \emph{(v) Simulators, Solvers \& Optimizers}, which evaluate actions, enforce constraints, or compute solutions in structured environments.

\subsection{Memory}
\label{sec:agent_memory}

Memory in time-series agents helps maintain coherence across reasoning steps, especially in temporally evolving settings~\cite{zhang2024surveymemorymechanismlarge}. We summarize memory by functional role: \emph{(i) Evidence logs}, which record intermediate traces of a decision process, such as intermediate predictions, retrieved evidence, tool outputs, candidate actions, and validation results; \emph{(ii) Pattern library}, which stores retrievable historical cases or typical data fragments, such as recurring market patterns, fault signatures, or similar past situations; and \emph{(iii) Analysis strategies}, which capture reusable experience about how to analyze a situation, such as which tools to use, which features to focus on, or how to interpret signals before acting.

%% file: 4_0_taxonomy.tex
\section{Taxonomy}
\label{sec:taxonomy}

We define an \emph{LLM agent for time series} as a system in which an LLM must make at least one decision that changes a multi-step time-series workflow, such as selecting an action or tool, updating memory, revising a hypothesis, or coordinating stages. We classify the system as single-agent when one LLM fills this role and as multi-agent when two or more LLMs take distinct decision-making roles and interact through cooperation, competition, or both. We exclude (i) one-shot prompting, (ii) pipelines in which the LLM serves only as a static encoder or post-hoc explainer, and (iii) domain-general agents that are not designed for time-series constraints or temporally grounded evaluation.

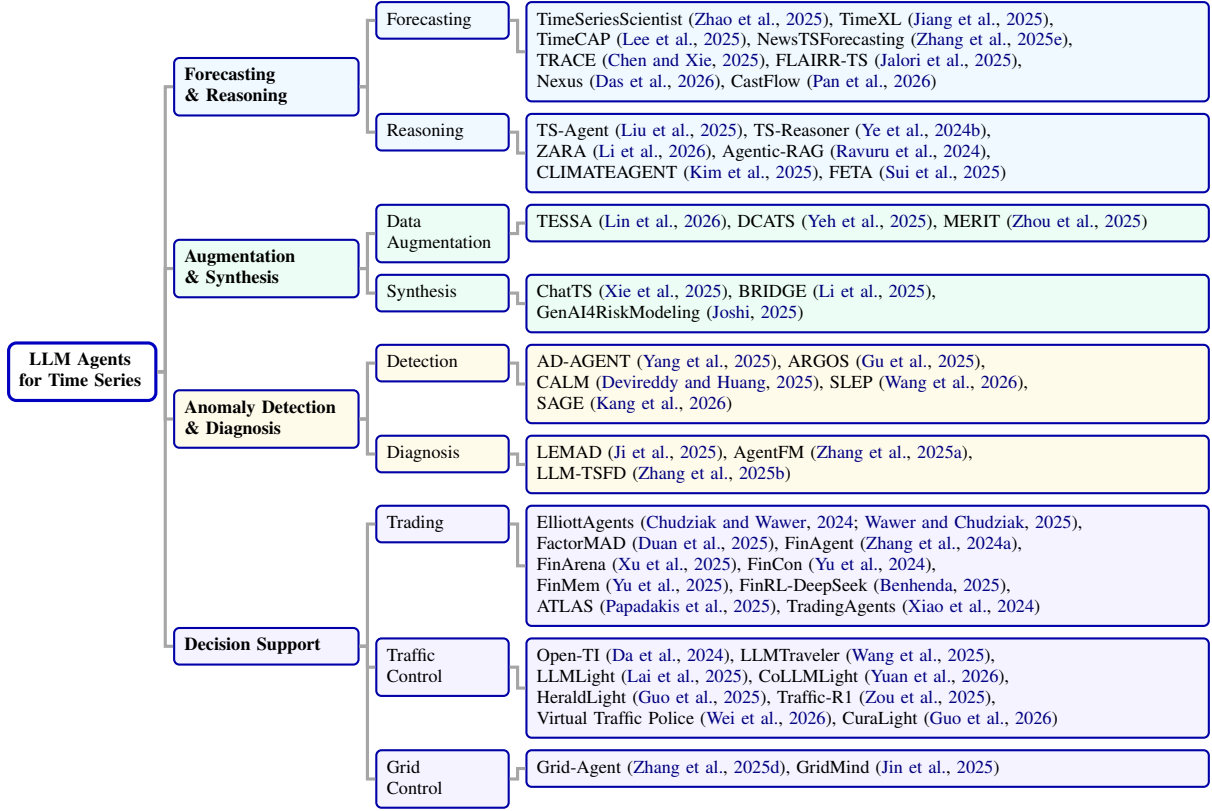
\begin{figure*}[t]
\centering
\resizebox{\textwidth}{!}{%
\begingroup
\emergencystretch=1em
\sloppy
\begin{forest} tsTax
[
  {\textbf{LLM Agents}\\ \textbf{for Time Series}}
  , tsRoot
  [
    {Forecasting\\ \& Reasoning}
    , tsMid, tsFamForecast, tsFamForecast
    [
      {Forecasting}
      , tsLeafNarrow
      [{TimeSeriesScientist~\cite{zhao2025timeseriesscientist}, TimeXL~\cite{jiang2025timexl}, \\TimeCAP~\cite{lee2025timecap}, NewsTSForecasting~\cite{zhang2025can}, \\TRACE~\cite{chen2025trace}, FLAIRR-TS~\cite{jalori2025flairr}, \\Nexus~\cite{das2026nexus}, CastFlow~\cite{pan2026castflow}}, tsLeaf]
    ]
    [
      {Reasoning}
      , tsLeafNarrow
      [{TS-Agent~\cite{liu2025ts}, TS-Reasoner~\cite{ye2024domain}, \\ZARA~\cite{li2025zara},  Agentic-RAG~\cite{ravuru2024agentic}, \\CLIMATEAGENT~\cite{kim2025climateagent}, FETA~\cite{sui2025feta}}, tsLeaf]
    ]
  ]
  [
    {Augmentation\\ \& Synthesis}
    , tsMid, tsFamData
    [
      {Data\\Augmentation}
      , tsLeafNarrow
      [{TESSA~\cite{lin2024decoding}, DCATS~\cite{yeh2025empowering}, MERIT~\cite{zhou2025merit}}, tsLeaf]
    ]
    [
      {Synthesis}
      , tsLeafNarrow
      [{ChatTS~\cite{xie2024chatts}, BRIDGE~\cite{li2025bridge}, \\GenAI4RiskModeling~\cite{joshi2025using}}, tsLeaf]
    ]
  ]
  [
    {Anomaly Detection\\ \& Diagnosis}
    , tsMid, tsFamAnomaly
    [
      {Detection}
      , tsLeafNarrow
      [{AD-AGENT~\cite{yang2025ad}, ARGOS~\cite{gu2501argos},\\ CALM~\cite{devireddy2025calm}, SLEP~\cite{wang2025large},\\ SAGE~\cite{kang2026sage}}, tsLeaf]
    ]
    [
      {Diagnosis}
      , tsLeafNarrow
      [{LEMAD~\cite{ji2025lemad}, AgentFM~\cite{zhang2025agentfm}, \\LLM-TSFD~\cite{zhang2025llm}}, tsLeaf]
    ]
  ]
  [
    {Decision Support}
    , tsMid, tsFamDecision
    [
      {Trading}
      , tsLeafNarrow
      [{ElliottAgents~\cite{chudziak2024elliottagents,wawer2025integrating}, \\FactorMAD~\cite{duan2025factormad},  FinAgent~\cite{zhang2024multimodal}, \\FinArena~\cite{xu2025finarena}, FinCon~\cite{yu2024fincon}, \\FinMem~\cite{yu2025finmem}, FinRL-DeepSeek~\cite{benhenda2025finrl}, \\ATLAS~\cite{papadakis2025atlas}, TradingAgents~\cite{xiao2024tradingagents}}, tsLeaf]
    ]
    [
      {Traffic\\Control}
      , tsLeafNarrow
      [{Open-TI~\cite{da2024open}, LLMTraveler~\cite{wang2025agentic}, \\LLMLight~\cite{lai2025llmlight}, CoLLMLight~\cite{yuan2025collmlight}, \\HeraldLight~\cite{guo2025heraldlight}, Traffic-R1~\cite{zou2025trafficr1}, \\Virtual Traffic Police~\cite{wei2026virtualtraffic}, CuraLight~\cite{guo2026curalight}}, tsLeaf]
    ]
    [
      {Grid\\Control}
      , tsLeafNarrow
      [{Grid-Agent~\cite{zhang2025grid}, GridMind~\cite{jin2025gridmind}}, tsLeaf]
    ]
  ]
]
\end{forest}
\endgroup
} 
\caption{A taxonomy of LLM agents for time-series tasks.}
\label{fig:taxonomy_overview}
\end{figure*}

We adopt a problem-driven taxonomy that groups systems by the time-series problems they address, and discuss design dimensions within each setting. This choice is user-oriented: readers are often more concerned with what kinds of designs are suitable for a given time-series problem than with design dimensions in isolation. A problem-driven view therefore provides clearer guidance for selecting and understanding agent designs in practice. Under this taxonomy, we identify four problem categories:
\emph{Forecasting \& Reasoning},
\emph{Augmentation \& Synthesis},
\emph{Anomaly Detection \& Diagnosis}, and
\emph{Decision Support}.
Within each category, we further distinguish several sub-problems.
Table~\ref{tab:taxonomy_overview} summarizes representative methods; Figure~\ref{fig:taxonomy_overview} shows the full taxonomy.

%% file: 4_1_taxonomy.tex
\subsection{Time-Series Forecasting \& Reasoning}
\label{subsec:forecasting_reasoning}

Time-series forecasting and reasoning share a common requirement: the agent must produce outputs that are not only plausible in language, but also grounded in explicit numerical or contextual evidence. In both settings, a fixed context window or a coarse global summary is often insufficient, because the correct conclusion may depend on local temporal patterns, historical analogs, reusable prototypes, or aligned external context such as news~\cite{zhang2025can}. As a result, effective systems usually do not treat context as a static input. Instead, they actively construct evidence beyond the context window, for example by retrieving relevant slices on demand~\cite{jalori2025flairr} or querying prototype memories~\cite{jiang2025timexl}.

\noindent\textbf{Time-Series Forecasting.} Time-series forecasting agents aim to predict future values from historical observations under non-stationarity, noise, and horizon-dependent uncertainty. In forecasting, agent systems are mainly shaped by: (i) multi-stage workflows, and (ii) competing hypotheses.

\emph{Multi-stage workflows} are central in forecasting because prediction is rarely a single-step task. Such workflows may involve data diagnosis, preprocessing, model selection, contextual analysis, prediction, and validation, so errors in early stages can invalidate later conclusions. A common design is therefore to ground each stage in explicit intermediate evidence rather than free-form reasoning alone. TimeSeriesScientist~\cite{zhao2025timeseriesscientist} is a representative example: it uses statistical models and other tools to generate diagnostics, validation results, and configuration records, and stores them as evidence logs for review, provenance, and correction. Nexus~\cite{das2026nexus} uses contextualization, dual-resolution macro/micro outlook generation, and synthesis/calibration stages, while CastFlow~\cite{pan2026castflow} organizes forecasting as planning, action, prediction, and reflection with memory and multi-view diagnostic tools.

\emph{Competing hypotheses} are another distinctive feature of forecasting. Different strategies may focus on different signals, temporal scales, or exogenous factors, so relying on a single reasoning path can be brittle. One design is to use competitive architectural patterns, where agents or candidate strategies represent forecasting views and are compared through explicit error feedback. NewsTSForecasting~\cite{zhang2025can} follows this idea by using error-based scoring and reflection to control strategy updates. TRACE~\cite{chen2025trace} similarly uses communication and multi-agent consistency refinement under sparse or missing observations, while also showing that consistency alone is not enough unless communication is checked against evidence.

\noindent\textbf{Time-Series Reasoning.} Time-series reasoning agents derive explanations, answers, or classifications from temporal evidence rather than directly predicting future values. In reasoning, two task-specific considerations are especially important: (i) domain knowledge and numerical support, and (ii) multi-step reasoning reliability.

\emph{Domain knowledge and numerical support} are essential in reasoning because text-trained LLMs do not reliably encode temporal structure, domain mechanisms, or quantitative operations. A common design is therefore to separate planning from computation: the main agent handles coordination, while numerical analysis is delegated to specialized sub-agents or auditable tools and operators. Agentic-RAG~\cite{ravuru2024agentic} illustrates the sub-agent route, while TS-Agent~\cite{liu2025ts} illustrates the tool-grounded route. Reasoning systems may also require domain knowledge beyond raw observations. ZARA~\cite{li2025zara} mines discriminative features offline and stores domain feature-importance profiles as reusable guidance, while CLIMATEAGENT~\cite{kim2025climateagent} uses specialized data agents to handle API conventions, metadata retrieval, and parameter validation. For classification-style reasoning, FETA~\cite{sui2025feta} decomposes multivariate series into channel-wise comparisons against retrieved exemplars and then aggregates confidence-weighted decisions.

\emph{Multi-step reasoning reliability} is similar to multi-stage forecasting workflows: reasoning tasks also involve a sequence of dependent steps, and early errors can propagate across the chain. A common design is therefore to make the process explicit, so intermediate results can be checked and revised rather than passed forward implicitly. CLIMATEAGENT~\cite{kim2025climateagent} follows the sub-agent route and records code, retrieved data, and results as evidence logs, so later stages can build on verified outputs. TS-Reasoner~\cite{ye2024domain} follows the operator route by compiling reasoning into an executable operator pipeline, where execution feedback can trigger plan revision and operator reselection. In both cases, evidence logs support downstream reasoning, reflection, and recovery. Verification may be handled by dedicated checking agents or by the same LLM in a critic role, as in TS-Agent~\cite{liu2025ts}.

\begin{tcolorbox}
[boxsep=0mm,left=2.5mm,right=2.5mm,boxrule=0.3pt,colframe=blue!60!black,colback=white]
\textbf{Discussion:} Forecasting and reasoning both need evidence beyond a fixed context window and are vulnerable to error accumulation in long workflows. These shared challenges motivate pattern libraries for retrieving precedents, evidence logs for preserving intermediate steps, and tool calls for quantitative support. The main difference is that forecasting more often compares competing hypotheses, whereas reasoning more often uses refinement loops to reduce long-chain errors.
\end{tcolorbox}

%% file: 4_2_taxonomy.tex
\subsection{Time-Series Augmentation \& Synthesis}
\label{subsec:augmentation_synthesis}

Time-series data augmentation and synthesis construct additional data while preserving time-series semantics. Their main failure mode is semantic drift: the constructed data may look plausible but violate the structures that matter for downstream tasks, such as trend, periodicity, local shapes, or correlations. The key difference is semantic source: augmentation mainly relies on the target series, whereas synthesis must align control text (e.g., scenario descriptions) with domain rules.

\noindent\textbf{Time-Series Data Augmentation.} Time-series data augmentation aims to construct additional data around a target sequence while preserving task-relevant semantics. It typically appears in two forms: generating numeric perturbations and generating textual annotations as an alternative representation of the same series. These two forms are shaped by different challenges: \emph{(i) semantic preservation} is central for numeric augmentation, while \emph{(ii) domain annotation understanding} is the main bottleneck for textual augmentation.

\emph{Semantic preservation} is the core challenge in numeric augmentation. Augmented data should remain aligned with the target series rather than merely look realistic in isolation. A common design follows one of two routes: either construct target-specific training sets from neighboring series, as in DCATS~\cite{yeh2025empowering}, or retrieve similar sequences and then apply classic augmentations such as jittering, scaling, and time warping, as in MERIT~\cite{zhou2025merit}. Verification is especially important here. LLM-as-Judge based only on pretrained knowledge can be brittle, so stronger designs usually rely on downstream validation signals~\cite{yeh2025empowering}.

\emph{Domain annotation understanding} is the main challenge in textual augmentation. In series-to-text semantic augmentation, the added data is not a new numeric sequence but a textual annotation, which can be viewed as another representation or semantic view of the same series. The difficulty is that pretrained LLMs do not reliably understand domain-specific time-series annotations. TESSA~\cite{lin2024decoding} derives domain-agnostic concepts (e.g., trend, periodicity, volatility) from cross-domain annotations and converts them into domain-specific annotations via a domain agent.

\noindent\textbf{Time-Series Synthesis.} Time-series synthesis aims to generate new sequences under specified controls or constraints rather than to expand a single target series. Under this view, it mainly includes two settings: text-guided synthesis, where the system generates sequences that satisfy both scenario descriptions and domain knowledge, and domain-attribute construction without external scenario text. These two settings face different bottlenecks: \emph{(i) text--series alignment} is central in the former, while \emph{(ii) domain constraint compliance} is the main challenge in the latter.

\emph{Text--series alignment} is the core challenge in text-guided synthesis. In real-world settings, paired scenario-query and time-series data are usually scarce, so LLMs cannot reliably map free-form text directly to numerical sequences. A common design is therefore to rewrite descriptions of real time series into structured text queries and then map these queries into executable generation settings. BRIDGE~\cite{li2025bridge} extracts templates (e.g., length, trend, periodicity, extrema, variance), refines them, and trains a controlled diffusion generator. GenAI4RiskModeling~\cite{joshi2025using} instead uses LLM-generated queries to tune GAN/VAE-based interest-rate scenario generation.

\emph{Domain constraint compliance} is the main challenge in domain-attribute construction without external scenario text. In this setting, the difficulty is not text alignment but ensuring that generated data still conforms to implicit domain rules and constraints. ChatTS~\cite{xie2024chatts} constructs synthetic time-series/text supervision by sampling domain-relevant attributes, generating rule-consistent series, and filtering Q\&A pairs for attribute consistency.

\begin{tcolorbox}
[boxsep=0mm,left=2.5mm,right=2.5mm,boxrule=0.3pt,colframe=green!45!black,colback=white]
\textbf{Discussion:} For augmentation and synthesis, the central issue is semantic drift: generated outputs may look plausible while violating task-relevant structures. Current systems therefore favor construction and verification pipelines, using data processing, generation modules, and downstream validation signals. Memory is less central here; when used, it mainly stores reusable templates or semantic patterns with domain rules.
\end{tcolorbox}

%% file: 4_3_taxonomy.tex
\subsection{Time-Series Anomaly Detection \& Diagnosis}
\label{subsec:ad_diagnosis}

Time-series anomaly detection and diagnosis are closely related: detection asks whether and when abnormal patterns occur, while diagnosis asks why and where they occur and how to respond. Their main difference lies in task emphasis.

\noindent\textbf{Time-Series Anomaly Detection.} In our scope, time-series anomaly detection focuses on producing reliable alarms, such as point-wise labels, anomalous windows, or alert events. In practice, three challenges are especially important: \emph{(i) evidence quality}, \emph{(ii) non-degradation} relative to the base detector, and \emph{(iii) streaming monitoring} under continual updates.

\emph{Evidence quality} matters because reliable alarms often require more than raw series values alone. Here, decision evidence refers to the information directly supporting the alarm decision, such as summary statistics, learned features, or retrieved historical snippets. A common design is to strengthen this evidence by injecting it into prompts or downstream modules. SLEP~\cite{wang2025large} follows this direction by enriching detector inputs with additional evidence rather than relying only on the raw sequence. SAGE~\cite{kang2026sage} makes this evidence construction more explicit by assigning specialized analyzers to point, structural, seasonal, and pattern anomalies, then consolidating their tool-grounded outputs into confidence-scored anomaly records.

\emph{Non-degradation} is critical because the agent is often layered on top of an existing deployed detector, so it should improve alarm quality without underperforming the base system. A common design is to treat the deployed detector as a base model and let the agent learn complementary corrections for its typical errors. ARGOS~\cite{gu2501argos} exemplifies this idea by constructing deterministic rules to correct base-detector failures and fusing rule outputs with detector outputs at inference time.

\emph{Streaming monitoring} is a special but practically important setting because concept drift may require continual adaptation. The main risk is contamination control: short-lived anomalies may be absorbed as the new normal during online updates. CALM~\cite{devireddy2025calm} addresses this with a continual design in which an LLM judge filters training data by distinguishing transient noise from sustained distribution shift, and only the latter is used to fine-tune the forecasting-based detector.

\noindent\textbf{Time-Series Diagnosis.} In our scope, time-series diagnosis focuses on explaining, localizing, and responding to abnormal events. In practice, two challenges are especially important: \emph{(i) evidence integration}, and \emph{(ii) limited diagnostic supervision}.

\emph{Evidence integration} is important for diagnosis for reasons similar to anomaly detection, but diagnosis places more emphasis on explanation and localization. In the papers we survey, this is often handled by explicitly incorporating textual evidence, such as logs, alerts, traces, and topology context, as part of the model input or generated report. AgentFM~\cite{zhang2025agentfm} follows this pattern and further improves stability through a RAG+CoT design that retrieves labeled historical examples as task-specific references, reducing free-form drift.

\emph{Limited diagnostic supervision} is another recurring bottleneck because labeled fault cases and high-quality incident narratives are often scarce. LLM-TSFD~\cite{zhang2025llm} addresses this with a human-in-the-loop data preparation stage, where users specify labeling or cleaning intent and the system generates executable code refined through feedback.

\begin{tcolorbox}
[boxsep=0mm,left=2.5mm,right=2.5mm,boxrule=0.3pt,colframe=orange!70!black,colback=white]
\textbf{Discussion:} Anomaly detection and diagnosis both build actionable monitoring pipelines from heterogeneous temporal evidence. This makes tool support central, especially data processing, detectors, analytical modules, and external evidence sources. Sequential cooperative pipelines remain common, while streaming scenarios emphasize memory updates and deployment-time adaptation. Memory mainly supports auditability and retrieval through evidence logs and historical pattern libraries.
\end{tcolorbox}

%% file: 4_4_taxonomy.tex
\subsection{Time-Series Decision Support}
\label{subsec:decision_making}

Time-series decision support differs from forecasting or diagnosis because the output is an executable action, plan, or allocation rather than a descriptive judgment. Decisions must be grounded in a predefined action space and satisfy domain constraints. Existing agentic work that satisfies our time-series scope is concentrated in trading, traffic control, and grid control; because the latter two share closed-loop control constraints, we discuss them together. We exclude decision-support agents in domains such as healthcare treatment and supply-chain operations when they do not directly use time-series evidence to produce executable decisions.

\noindent\textbf{Trading.} Trading agents focus on sequential financial decisions (e.g., buy, sell, hold) under evolving market conditions. Three considerations are central: (i) heterogeneous external evidence, (ii) historical experience, and (iii) risk preference.

\emph{Heterogeneous external evidence}, such as news, financial statements, social media, earnings calls, visual charts, and technical indicators, often affects trading decisions, but it comes in different forms and operates at different temporal scales. Processing all of it with a single agent can easily lead to context overload and mixed signals. A common cooperative pattern is a planner--executor structure, where specialized subagents or modules handle different evidence sources and their outputs are then aggregated. TradingAgents~\cite{xiao2024tradingagents}, FinCon~\cite{yu2024fincon}, and FinArena~\cite{xu2025finarena} all follow this pattern.

Trading decisions often rely heavily on \emph{historical experience}. In agent systems, this is usually supported by memory in the form of \emph{pattern library} and \emph{analytical strategies}, which abstract past situations into retrievable patterns or reusable decision experience. FinAgent~\cite{zhang2024multimodal} exemplifies this design: each stage produces a retrieval-oriented query, allowing market situations, price-driving explanations, and trading lessons to be stored separately. FinCon~\cite{yu2024fincon} further updates \emph{manager-level investment beliefs}, which serve as evolving analytical strategies.

\emph{Risk preference} is another special concern in trading. In some settings, the system needs to account for human preference alignment; FinArena~\cite{xu2025finarena} injects user risk preferences and feedback into prompts so that they directly influence the final recommendation. In other settings, the system does not explicitly incorporate user feedback, but instead constructs internal role-based variation to induce different risk styles, as in TradingAgents~\cite{xiao2024tradingagents} and FinMem~\cite{yu2025finmem}.
Recent evaluation work further emphasizes that static financial QA is insufficient for evaluating trading agents: StockBench~\cite{chen2025stockbench} evaluates agents in multi-month markets where daily prices, fundamentals, and news lead to sequential buy--sell--hold decisions.

\noindent\textbf{Traffic \& Grid Control.} Traffic and grid control agents both choose executable control actions from evolving infrastructure states. Traffic systems focus on signal actions under real-time flow constraints and network-level coordination, while grid systems target mitigation plans under changing loads, violations, contingencies, and uncertainty.

\emph{Hard constraints} are central because infrastructure actions must be evaluated in closed-loop environments. In traffic control, queue length, waiting time, throughput, and travel time evolve after each signal action. LLMLight~\cite{lai2025llmlight} operates in a fixed control space and is evaluated in a traffic simulator. More recent systems add stronger coordination and validation mechanisms: CoLLMLight~\cite{yuan2025collmlight} constructs a spatiotemporal graph for network-wide coordination, HeraldLight~\cite{guo2025heraldlight} uses a dual-LLM design with herald-guided prompts for fine-grained signal control, and Traffic-R1~\cite{zou2025trafficr1} trains a lightweight reasoning model for real-time signal control. In grid control, candidate actions such as switching, curtailment, or dispatch must be checked by power-flow solvers or safety validators. Grid-Agent~\cite{zhang2025grid} is a representative example: the LLM proposes structured mitigation plans, while power-flow solvers and validation modules determine their performance. GridMind~\cite{jin2025gridmind} similarly uses LLM agents for power-system analysis, with solvers providing domain-grounded feedback.

\emph{Human-specified policies} are another distinctive feature. In some systems, humans specify policies or optimization settings, and the LLM agent mainly orchestrates downstream execution. Open-TI~\cite{da2024open} illustrates this pattern well: some tasks translate human-described policies into signal actions, while others let the user specify simulation settings or optimization techniques and have the LLM route the request to the appropriate tool chain. Virtual Traffic Police~\cite{wei2026virtualtraffic} follows a related augmentation strategy by using an LLM agent to adjust parameters of existing traffic controllers under unforeseen incidents, while CuraLight~\cite{guo2026curalight} uses debate-guided data curation to improve an LLM-centered signal controller.

\begin{tcolorbox}
[boxsep=0mm,left=2.5mm,right=2.5mm,boxrule=0.3pt,colframe=violet!65!black,colback=white]
\textbf{Discussion:} Different decision targets lead to different design principles across trading, traffic control, and grid control. Trading systems more often use cooperative planner--executor designs to decompose multimodal evidence, whereas traffic and grid control systems rely more on simulator-grounded execution--validation pipelines. The same contrast appears in memory and tool use: trading emphasizes historical experience and heterogeneous data, while traffic and grid control depend more on simulation and feasibility checks.
\end{tcolorbox}

%% file: 5_resources.tex
\section{Resources}
\label{sec:resources}

This section summarizes key resources for implementing and evaluating LLM agents for time-series tasks. We organize them into four categories and provide representative examples (Table~\ref{tab:resources}).

\noindent\textbf{Datasets and Repositories.} These resources provide the raw observations used for training and offline evaluation. Representative forecasting datasets include Electricity~\cite{lai2018modeling}, METR-LA~\cite{li2017diffusion}, and ETT~\cite{zhou2021informer}, while common anomaly datasets include SWaT~\cite{mathur2016swat} and SMAP/MSL~\cite{hundman2018detecting}. Monash TSF~\cite{godahewa2021monash} serves as a multi-domain repository.

\noindent\textbf{Benchmarks.} Benchmarks define comparable tasks and metrics across methods. Forecasting benchmarks include M4/M5~\cite{makridakis2018m4,makridakis2022m5} and MIRAI~\cite{ye2024mirai}, while anomaly suites include NAB~\cite{lavin2015evaluating} and TSB-UAD~\cite{paparrizos2022tsb}. TimeSeriesExam~\cite{cai2024timeseriesexam} extends evaluation toward reasoning.

\noindent\textbf{Interactive Environments.} These platforms enable closed-loop experiments with explicit state, action, and reward signals. Typical examples include Grid2Op~\cite{marot2021learning} for power systems, SUMO~\cite{behrisch2011sumo} for traffic control, and SocioDojo~\cite{cheng2024sociodojo} for trading.

\noindent\textbf{Toolkits.} Toolkits provide reusable pipelines, APIs, and baselines for reproducible development. Common options include StatsForecast~\cite{garza2022statsforecast}, NeuralForecast~\cite{challu2023nhits}, Sktime~\cite{loning2019sktime}, and Stable-Baselines3~\cite{raffin2021stable}.

%% file: 6_future.tex
\section{Future Work}
\label{sec:future_work}

This survey reviews LLM-based agentic systems for time-series problems through a problem-driven taxonomy of forecasting and reasoning, augmentation and synthesis, anomaly detection and diagnosis, and decision support. Building on this taxonomy, we highlight four open directions.

\noindent\textbf{Numerical Understanding and Domain Knowledge.} LLMs remain limited in understanding numerical signals and specialized domain knowledge~\cite{hung2023walking,ye2024domain}. Even with external tools, agents must still interpret tool outputs and connect them to domain-specific reasoning.

\noindent\textbf{Causal and Counterfactual Temporal Reasoning.} Many time-series agents still rely mainly on correlations, while causal discovery from time series remains assumption-sensitive~\cite{assaad2022survey}. Diagnosis and decision support require evidence-grounded reasoning about interventions, delayed effects, and counterfactual outcomes; otherwise agents may hallucinate causal or temporal claims~\cite{wang2023counterfactual}.

\noindent\textbf{Online Adaptation and Continual Improvement.} In anomaly detection and decision support, agents often need to improve after deployment. Yet most systems remain static or rely on memory updates, while parameter adaptation remains rare~\cite{jaglan2025continual,zheng2026lifelong}.

\noindent\textbf{Benchmarking Agent Workflows.} Current studies evaluate time-series agents mainly with task metrics, leaving workflow quality undermeasured. Future benchmarks should evaluate intermediate decisions, tool use, validation, and reflection alongside downstream performance~\cite{weng2026temporalbench,cheng2026atsf}.

%% file: appendix_01_survey.tex
\section{Survey Scope and Related Surveys}
\label{sec:appendix_survey_comparison}
This appendix describes how we selected papers and how our survey differs from related surveys on LLMs for time-series analysis.

\paragraph{Paper selection.}
We searched arXiv, ACL Anthology, ACM Digital Library, IEEE Xplore, DBLP, Semantic Scholar, Google Scholar, and references from related surveys, with the last update in May 2026.
Search terms paired agent concepts, including \emph{LLM agent}, \emph{multi-agent}, \emph{tool use}, \emph{memory}, and \emph{reflection}, with time-series tasks and domains such as forecasting, anomaly detection, diagnosis, trading, traffic control, and power grids.
We included papers in which an LLM controls at least one decision that shapes a multi-step time-series workflow, including action or tool selection, memory updating, hypothesis revision, or stage coordination. We excluded one-shot prompting, static LLM encoders, post-hoc explainers, and domain-general agents without time-series evaluation.
The final corpus contains 47 representative systems; Table~\ref{tab:taxonomy_overview} codes their agent design, and Table~\ref{tab:comparable-all-tasks} reports comparable metrics where available.

\paragraph{Comparison with related surveys.}
Table~\ref{tab:survey_comparison} compares survey scope, task coverage, taxonomy, and design guidance. For categorical labels, \emph{Yes} denotes that the named dimension is a primary and sustained focus, \emph{Partial} denotes agent coverage without a clear separation from prompt-only methods, and \emph{Limited} denotes localized rather than category-wide design discussion. Method-, topology-, capability-, and problem-driven taxonomies organize methods by LLM adaptation, reasoning structure, agent capability, and time-series problem family, respectively.

Topology-driven surveys characterize reasoning structures~\cite{chang2025survey}; our problem-driven taxonomy complements them by linking task settings to architecture, tools, and memory. We cover 47 representative systems under the inclusion rule above and report per-system annotations in Table~\ref{tab:taxonomy_overview}.

\begin{table*}[!t]
\centering
\small
\setlength{\tabcolsep}{4pt}
\caption{Comparison with related time-series LLM surveys. $\checkmark$: clearly covered; $\sim$: partially covered; $\times$: not covered.}
\label{tab:survey_comparison}
\resizebox{\textwidth}{!}{%
\begin{tabular}{c c c p{1.0cm} c c c c p{2.4cm} p{1.9cm}}
\toprule
\textbf{Survey} & \textbf{Year} & \textbf{TS-specific} & \textbf{\shortstack{LLM\\Agents}} & \textbf{\shortstack{Forecasting\\\& Reasoning}} & \textbf{\shortstack{Augmentation\\\& Synthesis}} & \textbf{\shortstack{Anomaly Detection\\\& Diagnosis}} & \textbf{\shortstack{Decision\\Support}} & \textbf{Taxonomy} & \textbf{\shortstack{Design\\Guidance}} \\
\midrule
\cite{jiang2024empowering} & 2024 & Yes & No & $\checkmark$ & $\sim$ & $\checkmark$ & $\sim$ & Method-driven & Limited \\
\cite{zhang2024large} & 2024 & Yes & No & $\checkmark$ & $\checkmark$ & $\checkmark$ & $\times$ & Method-driven & Limited \\
\cite{chang2025survey} & 2026 & Yes & Partial & $\checkmark$ & $\checkmark$ & $\checkmark$ & $\checkmark$ & Topology-driven & Limited \\
\cite{zhangprompts} & 2025 & Yes & Partial & $\checkmark$ & $\checkmark$ & $\checkmark$ & $\sim$ & Capability-driven & Limited \\
\textbf{Ours} & 2026 & Yes & Yes & $\checkmark$ & $\checkmark$ & $\checkmark$ & $\checkmark$ & Problem-driven & Yes \\
\bottomrule
\end{tabular}}
\end{table*}

%% file: appendix_03_prompts.tex
\ifshowappendixfloats\else
\section{Representative Prompt Examples by Task Family}
\label{sec:appendix_prompt_examples}

This appendix provides representative prompt examples for the major task
families in our taxonomy. The examples are abstracted from recurring
input--output needs in the surveyed systems. They are not verbatim prompts from
any single paper, and they are not intended to define a separate taxonomy of
prompt types.

\lstdefinestyle{promptstyle}{
  basicstyle=\ttfamily\scriptsize,
  breaklines=true,
  breakatwhitespace=false,
  columns=fullflexible,
  keepspaces=true,
  showstringspaces=false,
  upquote=true
}

\fi

\ifshowappendixfloats\else
\subsection{Forecasting and Reasoning}

\begin{tcblisting}{
  breakable,
  listing only,
  colback=famForecast,
  colframe=blue!60!black,
  boxrule=0.3pt,
  boxsep=1mm,
  left=1mm,
  right=1mm,
  listing options={style=promptstyle}
}
System:
You are a time-series analysis agent.
Use temporal evidence, retrieved context,
and tool outputs to produce a grounded
forecast, answer, or classification. Do
not rely on unsupported patterns.

User:
Task:
- Forecast the target horizon, or answer
  the temporal reasoning or classification
  question.

Inputs:
- Historical observations:
  <series or summarized windows>
- Target horizon or question:
  <forecast horizon / question>
- Retrieved analogs or external context:
  <optional>
- Candidate hypotheses or exemplars:
  <optional>
- Tool outputs:
  <trend, seasonality, anomaly,
   correlation, model scores>

Instructions:
1. Identify trend, seasonality, local
   changes, and unusual events.
2. Select evidence relevant to the
   horizon or question.
3. Compare candidate hypotheses or
   retrieved exemplars when provided.
4. Use tool outputs for quantitative
   claims.
5. Produce the forecast, answer, or label
   with a concise rationale.
6. State uncertainty and failure modes.

Output JSON:
{
	  "evidence_summary": "...",
	  "reasoning": "...",
	  "prediction_answer_or_label": "...",
	  "confidence": "...",
	  "limitations": "..."
}
\end{tcblisting}

This example reflects common input--output needs in forecasting and reasoning
agents such as TimeSeriesScientist~\cite{zhao2025timeseriesscientist},
TimeXL~\cite{jiang2025timexl}, TS-Agent~\cite{liu2025ts}, and
TS-Reasoner~\cite{ye2024domain}.

\subsection{Augmentation and Synthesis}

\begin{tcblisting}{
  breakable,
  listing only,
  colback=famData,
  colframe=green!45!black,
  boxrule=0.3pt,
  boxsep=1mm,
  left=1mm,
  right=1mm,
  listing options={style=promptstyle}
}
System:
You are assisting time-series data
construction. Augment a target series or
synthesize a controlled series only when
the requested temporal semantics and
domain constraints can be preserved.

User:
Task:
- Create an augmented example, synthetic
  series, or textual annotation.

Inputs:
- Construction mode:
  <target augmentation | text-guided
   synthesis | domain-attribute synthesis>
- Seed series or source examples:
  <values / windows / examples>
- Desired label, scenario, or query:
  <class, event, scenario, or description>
- Required temporal properties:
  <trend, periodicity, volatility, extrema>
- Domain constraints:
  <valid ranges, correlations,
   physical rules>
- Validation criteria:
  <downstream check or consistency rule>

Instructions:
1. Preserve task-relevant temporal
   semantics when augmenting a target
   series.
2. Align control text or domain attributes
   with the generated series when
   synthesizing new data.
3. Apply only transformations consistent
   with the label or scenario.
4. Avoid changing causal, seasonal, or
   domain-critical structure.
5. Explain which properties are preserved
   or intentionally changed.
6. Return a compact structured result for
   downstream validation.

Output JSON:
{
  "construction_mode": "...",
	  "constructed_item": "...",
	  "control_attributes": "...",
	  "preserved_properties": "...",
  "changed_properties": "...",
  "validation_notes": "..."
}
\end{tcblisting}

This example reflects common input--output needs in augmentation and synthesis
systems such as TESSA~\cite{lin2024decoding}, MERIT~\cite{zhou2025merit},
BRIDGE~\cite{li2025bridge}, and ChatTS~\cite{xie2024chatts}.

\subsection{Anomaly Detection and Diagnosis}

\begin{tcblisting}{
  breakable,
  listing only,
  colback=famAnomaly,
  colframe=orange!70!black,
  boxrule=0.3pt,
  boxsep=1mm,
  left=1mm,
  right=1mm,
  listing options={style=promptstyle}
}
System:
You are a time-series monitoring agent.
Convert detector evidence, temporal
context, and auxiliary logs into a
supported alarm, point/window label, or
diagnostic explanation.

User:
Task:
- Decide whether the candidate window is
  anomalous, or explain a fault.

Inputs:
- Candidate window:
  <time range and observed values>
- Detector evidence:
  <scores, thresholds, labels, residuals>
- Temporal context:
  <recent history, seasonality,
   expected behavior>
- Auxiliary context:
  <logs, alerts, topology,
   related variables>
- Historical cases:
  <optional retrieved examples>

Instructions:
1. Compare the candidate window with
   expected temporal behavior.
2. Separate detector evidence from
   contextual or textual evidence.
3. Check whether context supports or
   contradicts the detector evidence.
4. Identify the most likely abnormal
   interval or root cause.
5. Avoid overriding the detector or
   overclaiming when evidence is
   weak or conflicting.
6. Return a decision, supporting evidence,
   and confidence.

Output JSON:
{
  "decision": "normal | anomalous | uncertain",
  "abnormal_interval": "...",
  "supporting_evidence": "...",
  "root_cause_hypothesis": "...",
  "confidence": "..."
}
\end{tcblisting}

This example reflects common input--output needs in anomaly detection and
diagnosis systems such as SAGE~\cite{kang2026sage}, ARGOS~\cite{gu2501argos},
AgentFM~\cite{zhang2025agentfm}, and LLM-TSFD~\cite{zhang2025llm}.

\subsection{Decision Support}

\begin{tcblisting}{
  breakable,
  listing only,
  colback=famDecision,
  colframe=violet!65!black,
  boxrule=0.3pt,
  boxsep=1mm,
  left=1mm,
  right=1mm,
  listing options={style=promptstyle}
}
System:
You are a sequential decision-support
agent. Recommend only actions allowed by
the action space and supported by the
current temporal state, constraints, and
validation feedback.

User:
Task:
- Select the next trading, traffic-control,
  or grid-control action.

Inputs:
- Current state:
  <market / intersection / grid state>
- Recent history:
  <prices, flows, loads, events,
   or violations>
- Historical experience:
  <retrieved cases, lessons, or patterns>
- External evidence:
  <news, indicators, forecasts,
   alerts, optional>
- Action space:
  <allowed actions and parameter ranges>
- Constraints:
  <risk preference, safety limits,
   policies, timing, feasibility>
- Validation feedback:
  <simulator, solver, or previous outcome>

Instructions:
1. Summarize the state variables that
   matter for the next action.
2. Compare feasible actions under the
   provided constraints.
3. Use historical experience for
   market-like settings when available.
4. Use validation feedback to reject
   unsafe or dominated actions.
5. Recommend one action and explain the
   expected effect.
6. State the main risk and what should be
   monitored next.

Output JSON:
{
  "state_summary": "...",
  "recommended_action": "...",
  "justification": "...",
  "constraint_check": "...",
  "next_monitoring_target": "..."
}
\end{tcblisting}

This example reflects common input--output needs in decision-support systems
such as TradingAgents~\cite{xiao2024tradingagents}, FinCon~\cite{yu2024fincon},
LLMLight~\cite{lai2025llmlight}, and Grid-Agent~\cite{zhang2025grid}.
\fi

%% file: appendix_04_patterns.tex
\ifshowappendixfloats\else
\section{Typical Design Patterns}
\label{sec:appendix_patterns}

Figure~\ref{fig:appendix_patterns} provides a supplementary visual summary of the typical design patterns discussed across different time-series tasks in Section~\ref{sec:taxonomy}. Each panel shows a common design pattern rather than the exact design of a specific system. Dashed boxes denote agent modules together with their tools, solid boxes group entities with similar roles, solid arrows show the main flow, dashed arrows indicate conditional or fallback paths, and background colors group panels from the same higher-level problem type.
\fi

\ifshowappendixfloats
\begin{figure*}[p]
  \centering
  \includegraphics[width=\textwidth]{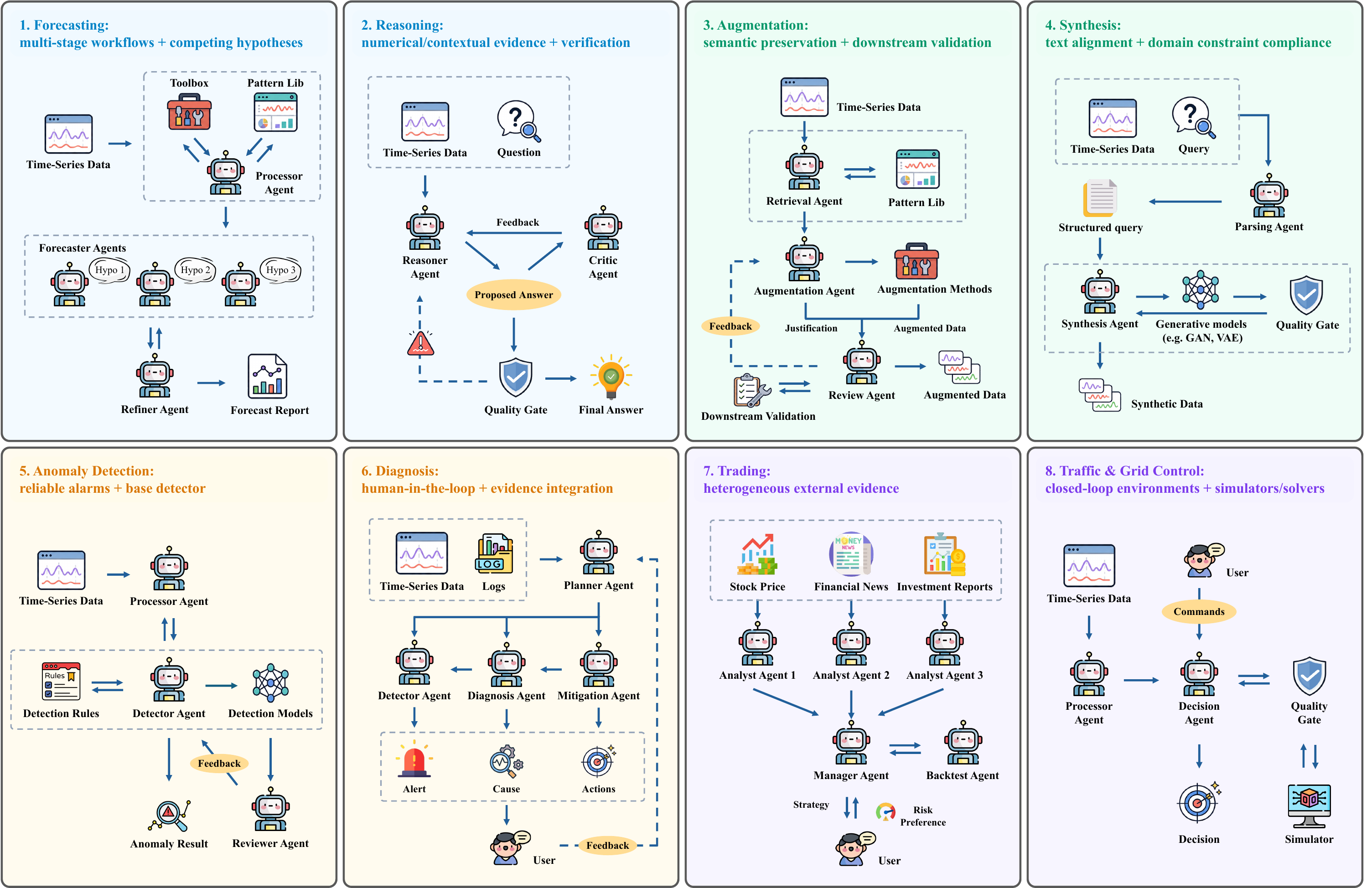}
  \caption{Typical design patterns of time-series agent systems across representative tasks. From top left to bottom right, the panels correspond to forecasting, reasoning, augmentation, synthesis, anomaly detection, diagnosis, trading, and traffic and grid control.}
  \label{fig:appendix_patterns}
\end{figure*}
\fi

%% file: appendix_07_failure_modes.tex
\section{Failure Modes and Design Patterns}
\label{sec:appendix_failure_modes}

The task sections discuss these failure modes in context. Table~\ref{tab:failure_modes_design_patterns} brings them into one view and links each one to design choices reported in related work. The table is not a causal comparison; it is meant as a practical checklist for where a design choice helps and what should be inspected.

\begin{table*}[!t]
\centering
\scriptsize
\setlength{\tabcolsep}{2.5pt}
\renewcommand{\arraystretch}{1.16}
\caption{Summary of documented failure modes for time-series agent systems. Each failure mode is linked to related work, design choices, and implementation checks.}
\label{tab:failure_modes_design_patterns}
\resizebox{\textwidth}{!}{%
\begin{tabular}{p{2.45cm} p{3.0cm} p{3.4cm} p{4.6cm} p{3.0cm}}
\toprule
\textbf{Failure mode} & \textbf{Where it appears} & \textbf{Related work} & \textbf{Design choice} & \textbf{Implementation check} \\
\midrule
Error propagation across workflow stages & Forecasting and reasoning pipelines that pass outputs through diagnosis, model selection, context analysis, prediction, and validation & TimeSeriesScientist~\cite{zhao2025timeseriesscientist}; CLIMATEAGENT~\cite{kim2025climateagent}; CastFlow~\cite{pan2026castflow} & Keep stage outputs explicit. Log retrieved evidence, tool calls, predictions, and validation results so later stages do not inherit hidden errors & Can a reader inspect the artifact used by the next stage? \\
\addlinespace
Numerical or domain reasoning errors & Tasks that require temporal structure, metadata, domain conventions, or quantitative operations & TS-Agent~\cite{liu2025ts}; TS-Reasoner~\cite{ye2024domain}; CLIMATEAGENT~\cite{kim2025climateagent} & Let the LLM plan and explain, but move calculation and domain-specific parsing into tools, operators, or specialist agents & Can the numerical claim be rerun outside the LLM? \\
\addlinespace
Semantic drift in generated data & Augmentation and synthesis, where plausible-looking outputs can break trend, periodicity, local shape, correlation, or domain constraints & DCATS~\cite{yeh2025empowering}; MERIT~\cite{zhou2025merit}; BRIDGE~\cite{li2025bridge} & Generate from target context and verify temporal properties before the synthetic data enters training or evaluation & Do checks cover trend, seasonality, local shape, and correlation rather than textual plausibility alone? \\
\addlinespace
Brittle LLM-as-judge validation & Validation settings where a pretrained judge can approve an output without task evidence or executable checks & DCATS~\cite{yeh2025empowering}; TS-Reasoner~\cite{ye2024domain} & Treat LLM critique as one signal. Require task metrics, execution feedback, or downstream validation before accepting an output & Would the output be rejected if another LLM approved it but the executable check failed? \\
\addlinespace
Contamination during online updates & Streaming anomaly detection, where transient anomalies can be absorbed as the new normal & CALM~\cite{devireddy2025calm} & Separate alarm decisions from update decisions; gate which observations can enter memory or fine-tuning data & Can short-lived anomalies be kept out of the update set? \\
\addlinespace
Unsafe or infeasible actions in closed-loop control & Traffic and grid-control agents that issue executable actions under operational constraints & LLMLight~\cite{lai2025llmlight}; Grid-Agent~\cite{zhang2025grid}; GridMind~\cite{jin2025gridmind} & Check proposed actions with a simulator, solver, feasibility test, or safety validator before execution & Does every action pass an environment or constraint check first? \\
\bottomrule
\end{tabular}}
\end{table*}

%% file: appendix_05_resources.tex
\ifshowappendixfloats\else
\section{Resources Summary}
\label{sec:appendix_table2}

Table~\ref{tab:resources} summarizes the resources discussed in Section~\ref{sec:resources}, including benchmarks, datasets, environments, and toolkits. For each resource, we report its category, associated problem type, interactivity, temporal scale, number of series, and latest release when available.
\fi

\ifshowappendixfloats
\begin{table*}[p]
\centering
\caption{Summary of representative resources for time-series tasks. Each resource is grouped by \textbf{Problem Type} and further characterized by its type, interactivity, temporal scale, number of series, and latest release.}
\label{tab:resources}
\scriptsize
\resizebox{\textwidth}{!}{\begin{tabular}{p{0.23\linewidth} p{0.12\linewidth} p{0.16\linewidth} p{0.07\linewidth} p{0.08\linewidth} p{0.08\linewidth} p{0.08\linewidth}}
\toprule
\textbf{Resource} & \textbf{Category} & \textbf{Problem Type} & \textbf{Interactive} & \textbf{Timesteps} & \textbf{Series} & \textbf{Latest Release} \\
\midrule

\multicolumn{7}{l}{\textit{Forecasting}} \\

\href{https://www.unic.ac.cy/iff/research/forecasting/m-competitions/m4/}{M4} \cite{makridakis2018m4} & Benchmark & Forecasting & No & Up to 10k & $\sim$100k & 2018 \\
\href{https://www.unic.ac.cy/iff/research/forecasting/m-competitions/m5/}{M5} \cite{makridakis2022m5} & Benchmark & Forecasting & No & 1.9k & 42k & 2020 \\
\href{https://mirai-llm.github.io/}{MIRAI} \cite{ye2024mirai} & Benchmark & Forecasting & No & $\sim$ 1M & 59k & 2025\\
\href{https://futurex-ai.github.io/}{FutureX} \cite{zeng2025futurex} & Benchmark & Forecasting & No & Live/Daily & 195 sources & 2025 \\
\href{https://forecastingdata.org/}{Monash TSF} \cite{godahewa2021monash} & Dataset Repo & Multi-Domain Forecasting & No & Up to 527k & Up to 145k & 2021 \\

\href{https://archive.ics.uci.edu/ml/datasets/individual+household+electric+power+consumption}{IHEPC} & Dataset & Energy Forecasting & No & 2M & 9 & 2011 \\
\href{https://archive.ics.uci.edu/dataset/321/electricityloaddiagrams20112014}{Electricity} \cite{lai2018modeling} & Dataset & Energy Forecasting & No & 26k & 321 & 2015 \\
\href{https://github.com/liyaguang/DCRNN}{METR-LA} \cite{li2017diffusion} & Dataset & Traffic Forecasting & No & 34k & 207 & 2018 \\
\href{https://github.com/liyaguang/DCRNN}{PEMS-BAY} \cite{li2017diffusion} & Dataset & Traffic Forecasting & No & 52k & 325 & 2018 \\
\href{https://github.com/zhouhaoyi/ETDataset}{ETT} \cite{zhou2021informer} & Dataset & Energy Forecasting & No & Up to 69k & 7 & 2021 \\
\href{https://github.com/laiguokun/multivariate-time-series-data}{Exchange} \cite{lai2018modeling} & Dataset & Finance Forecasting & No & 7.6k & 8 & 2021 \\
\href{https://github.com/laiguokun/multivariate-time-series-data}{Traffic} \cite{lai2018modeling} & Dataset & Traffic Forecasting & No & 17k & 862 & 2023 \\
\href{https://nixtlaverse.nixtla.io/datasetsforecast/long_horizon2.html}{Weather} & Dataset & Weather Forecasting & No & 53k & 21 & 2023 \\
\href{https://github.com/liuxu77/LargeST}{LargeST} \cite{liu2023largest} & Dataset & Traffic Forecasting & No & 526k & Up to 8.6k & 2023 \\
\href{https://www.cdc.gov/fluview/index.html}{ILI} & Dataset & Health Forecasting & No & Unclear & 7 & 2026 \\
\href{https://github.com/Nixtla/statsforecast}{StatsForecast} \cite{garza2022statsforecast} & Toolkit & Forecasting & No & N/A & N/A & 2025 \\
\href{https://github.com/Nixtla/neuralforecast}{NeuralForecast} \cite{challu2023nhits} & Toolkit & Forecasting & No & N/A & N/A & 2026 \\

\midrule
\multicolumn{7}{l}{\textit{Reasoning}} \\

\href{https://github.com/moment-timeseries-foundation-model/TimeSeriesExam}{TimeSeriesExam} \cite{cai2024timeseriesexam} & Benchmark & Reasoning & No & N/A & >700 Tasks & 2024 \\
\href{https://www.ncei.noaa.gov/products/international-best-track-archive}{IBTrACS} \cite{knapp2010international}& Dataset & Reasoning & No & N/A & Unclear & 2025 \\

\midrule
\multicolumn{7}{l}{\textit{Anomaly Detection}} \\

\href{https://github.com/TheDatumOrg/TSB-UAD}{TSB-UAD} \cite{paparrizos2022tsb} & Benchmark Suite & Anomaly Detection & No & Varies & 12.7k TS & 2022 \\
\href{https://github.com/Minqi824/ADBench}{ADBench} \cite{han2022adbench} & Benchmark Suite & Anomaly Detection & No & Varies & 57 Datasets & 2022 \\
\href{https://github.com/numenta/NAB}{NAB} \cite{lavin2015evaluating} & Benchmark & Anomaly Detection & No & Up to 22k & 58 & 2015 \\
\href{https://wu.renjie.im/research/anomaly-benchmarks-are-flawed/}{UCR-AD} \cite{wu2021current} & Benchmark & Anomaly Detection & No & Unclear & 250 & 2021 \\
\href{https://itrust.sutd.edu.sg/itrust-labs_datasets/}{SWaT} \cite{mathur2016swat} & Dataset & Anomaly Detection & No & 11 days & 51 & 2016 \\
\href{https://github.com/khundman/telemanom}{SMAP} \cite{hundman2018detecting} & Dataset & Anomaly Detection & No & 430k & 55 & 2018 \\
\href{https://github.com/khundman/telemanom}{MSL} \cite{hundman2018detecting} & Dataset & Anomaly Detection & No & 67k & 27 & 2018 \\
\href{https://itrust.sutd.edu.sg/itrust-labs_datasets/}{WADI} \cite{adepu2020investigation} & Dataset & Anomaly Detection & No & Unclear & 103 & 2019 \\
\href{https://github.com/NetManAIOps/KPI-Anomaly-Detection}{KPI} \cite{ren2019time} & Dataset & Anomaly Detection & No & Unclear & $\sim$29 KPIs & 2019 \\
\href{https://github.com/arundo/adtk}{ADTK} & Toolkit & Anomaly Detection & No & N/A & N/A & 2020 \\
\href{https://github.com/linkedin/luminol}{Luminol} & Toolkit & Anomaly Detection & No & N/A & N/A & 2017 \\

\midrule
\multicolumn{7}{l}{\textit{Decision Support}} \\

\href{https://github.com/deepmind/deepmind-research/tree/master/rl_unplugged}{RL Unplugged} \cite{gulcehre2020rl} & Dataset & Offline RL & No & N/A & N/A & 2020 \\
\href{https://github.com/Farama-Foundation/D4RL}{D4RL} \cite{fu2020d4rl} & Dataset & Offline RL & No & N/A & N/A & 2021 \\

\href{https://github.com/Grid2op/grid2op}{Grid2Op} \cite{marot2021learning} & Environment & Power System Control & Yes & N/A & N/A & 2026 \\
\href{https://github.com/chengjunyan1/SocioDojo}{SocioDojo} \cite{cheng2024sociodojo} & Environment & Trading & Yes & N/A & N/A & 2024 \\
\href{https://www.citylearn.net/citylearn_challenge/2021.html}{CityLearn} & Environment & Building Energy Control & Yes & N/A & N/A & 2025 \\
\href{https://eclipse.dev/sumo/}{SUMO} \cite{behrisch2011sumo} & Environment & Traffic Control & Yes & N/A & N/A & 2026 \\
\href{https://github.com/DLR-RM/stable-baselines3}{Stable-Baselines3} \cite{raffin2021stable} & Toolkit & Reliable RL & No & N/A & N/A & 2026 \\

\midrule
\multicolumn{7}{l}{\textit{Others}} \\

\href{https://github.com/moment-timeseries-foundation-model/TimeSeriesGym}{TimeSeriesGym} \cite{cai2025timeseriesgym} & Benchmark & Multi-Task & No & N/A & N/A & 2025 \\
\href{https://github.com/sktime/sktime}{Sktime} \cite{loning2019sktime} & Toolkit & Multi-Task & No & N/A & N/A & 2025 \\
\href{https://github.com/awslabs/gluonts}{GluonTS} \cite{alexandrov2020gluonts} & Toolkit & Multi-Task & No & N/A & N/A & 2025 \\
\href{https://github.com/salesforce/Merlion}{Merlion} \cite{bhatnagar2021merlion} & Toolkit & Multi-Task & No & N/A & N/A & 2024 \\

\bottomrule
\end{tabular}}
\end{table*}
\fi

%% file: appendix_06_evaluation.tex
\ifshowappendixfloats\else
\section{Evaluation Evidence and Metrics}
\label{app:evaluation}

Table~\ref{tab:comparable-all-tasks} reports representative comparable quantitative evidence across forecasting, reasoning, anomaly detection, trading, and traffic-control tasks.
We only group results that share the same task, dataset, and metric, and we mark the surveyed agent methods in bold.
The table is intended as comparable evidence rather than a universal leaderboard: scores may still depend on horizon, split, point-adjustment rule, transaction-cost assumption, or simulator configuration.

\paragraph{Error metrics.}
Regression-style tasks usually report point-wise errors such as mean absolute error (MAE), mean squared error (MSE), and root mean squared error (RMSE):
\[
\begin{aligned}
\mathrm{MAE} &= \frac{1}{n}\sum_{i=1}^{n}|y_i-\hat{y}_i|,\\
\mathrm{MSE} &= \frac{1}{n}\sum_{i=1}^{n}(y_i-\hat{y}_i)^2,\\
\mathrm{RMSE} &= \sqrt{\mathrm{MSE}}.
\end{aligned}
\]
Here $y_i$ and $\hat{y}_i$ denote the target and prediction for the $i$-th evaluated point, and $n$ is the number of evaluated points. Lower values indicate better forecasts. Scale-free variants such as MAPE, sMAPE, or MASE are used when series with different magnitudes must be compared.

\paragraph{Classification and detection metrics.}
Reasoning, event prediction, anomaly detection, and some augmentation studies are commonly evaluated with accuracy, precision, recall, and F1. Let TP, FP, and FN denote true positives, false positives, and false negatives:
\[
\begin{aligned}
\mathrm{Precision} &= \frac{\mathrm{TP}}{\mathrm{TP}+\mathrm{FP}}, \quad
\mathrm{Recall} = \frac{\mathrm{TP}}{\mathrm{TP}+\mathrm{FN}}, \\
\mathrm{F1} &= 2\frac{\mathrm{Precision}\cdot\mathrm{Recall}}
{\mathrm{Precision}+\mathrm{Recall}}.
\end{aligned}
\]
Macro-F1 averages class-wise F1 and is useful under class imbalance; micro-F1 aggregates counts before computing F1. In anomaly detection, F1 can be point-wise, point-adjusted, or event-based, so comparisons require the same adjustment rule. Score-based detectors may additionally use AUROC or AUPR.

\paragraph{Generation and annotation metrics.}
For augmentation and synthesis, evaluation is often indirect: generated samples are used to train or adapt a downstream model, and the downstream MAE, MSE, accuracy, or F1 is reported. Papers may also measure distributional fidelity with DTW, MMD, autocorrelation, spectral statistics, or nearest-neighbor analyses, and annotation quality with expert agreement or label accuracy.

\paragraph{Decision and control metrics.}
Trading agents are evaluated by both return and risk. Cumulative return (CR) measures portfolio growth, while the Sharpe ratio (SR) measures risk-adjusted return:
\[
\mathrm{SR}=\frac{\mathbb{E}[R_t-R_f]}{\sigma(R_t-R_f)}.
\]
Maximum drawdown, volatility, and turnover further capture downside risk and trading cost sensitivity. Recent trading-agent benchmarks such as StockBench~\cite{chen2025stockbench} further stress multi-month sequential evaluation under daily market signals.

Traffic-control agents are usually evaluated in closed-loop simulators. Average travel time (ATT) is the mean travel duration across vehicles:
\[
\mathrm{ATT}=\frac{1}{N}\sum_{i=1}^{N}(t_i^{\mathrm{exit}}-t_i^{\mathrm{entry}}),
\]
where lower ATT indicates more efficient traffic flow. Related metrics include waiting time, queue length, throughput, cumulative reward, and constraint violations.
\fi

\ifshowappendixfloats
\begin{table*}[p]
\centering
\caption{Comparable quantitative evidence across time-series tasks (Part 1). Agent methods are highlighted in bold.}
\label{tab:comparable-all-tasks}
\scriptsize
\renewcommand{\arraystretch}{0.86}
\setlength{\tabcolsep}{3pt}
\resizebox{\textwidth}{!}{%
\begin{tabular}{p{0.16\linewidth} p{0.16\linewidth} p{0.11\linewidth} p{0.34\linewidth} p{0.08\linewidth}}
\toprule
\textbf{Task} & \textbf{Dataset} & \textbf{Metric} & \textbf{Method} & \textbf{Score} \\
\midrule

\multicolumn{5}{l}{\textit{ETTh1 forecasting} \cite{zhou2021informer}} \\
Forecasting & ETTh1 & MAE & \textbf{FLAIRR-TS} \cite{jalori2025flairr} & 0.101 \\
Forecasting & ETTh1 & MAE & LSTP \cite{liu2024lstprompt} & 0.150 \\
Forecasting & ETTh1 & MAE & DLinear \cite{zeng2023transformers} & 0.390 \\
Forecasting & ETTh1 & MAE & \textbf{Agentic-RAG (Llama3-8B)} \cite{ravuru2024agentic} & 0.396 \\
Forecasting & ETTh1 & MAE & GPT4TS \cite{zhou2023onefitsall} & 0.397 \\
Forecasting & ETTh1 & MAE & PatchTST \cite{nie2023time} & 0.399 \\
Forecasting & ETTh1 & MAE & TimesNet \cite{wu2023timesnet} & 0.402 \\
Forecasting & ETTh1 & MAE & FEDFormer \cite{zhou2022fedformer} & 0.419 \\
Forecasting & ETTh1 & MAE & Time-LLM \cite{jin2023timellm} & 0.460 \\

\midrule
\multicolumn{5}{l}{\textit{Weather forecasting} \cite{lee2025timecap}} \\
Forecasting & Weather & F1 & \textbf{TimeXL (GPT-4o)} \cite{jiang2025timexl} & 0.696 \\
Forecasting & Weather & F1 & \textbf{TimeCAP (GPT-4)} \cite{lee2025timecap} & 0.668 \\
Forecasting & Weather & F1 & FreTS \cite{yi2024frequency} & 0.623 \\
Forecasting & Weather & F1 & Time-LLM \cite{jin2023timellm} & 0.613 \\
Forecasting & Weather & F1 & PatchTST \cite{nie2023time} & 0.592 \\
Forecasting & Weather & F1 & LLMTime \cite{gruver2023large} & 0.587 \\
Forecasting & Weather & F1 & Autoformer \cite{wu2021autoformer} & 0.546 \\
Forecasting & Weather & F1 & iTransformer \cite{liu2024itransformer} & 0.541 \\
Forecasting & Weather & F1 & DLinear \cite{zeng2023transformers} & 0.540 \\
Forecasting & Weather & F1 & Crossformer \cite{zhang2023crossformer} & 0.500 \\
Forecasting & Weather & F1 & PromptCast \cite{xue2023promptcast} & 0.499 \\
Forecasting & Weather & F1 & TimesNet \cite{wu2023timesnet} & 0.494 \\
Forecasting & Weather & F1 & TSMixer \cite{chen2023tsmixer} & 0.488 \\

\midrule
\multicolumn{5}{l}{\textit{TimeSeriesExam reasoning} \cite{cai2024timeseriesexam}} \\
Reasoning & TimeSeriesExam & Accuracy & \textbf{TS-Agent (GPT-4o-mini)} \cite{liu2025ts} & 0.55 \\
Reasoning & TimeSeriesExam & Accuracy & GPT-o1 & 0.37 \\
Reasoning & TimeSeriesExam & Accuracy & GPT-4o & 0.29 \\
Reasoning & TimeSeriesExam & Accuracy & DeepSeek & 0.28 \\
Reasoning & TimeSeriesExam & Accuracy & Phi-3.5 & 0.25 \\

\midrule
\multicolumn{5}{l}{\textit{HAR average reasoning} \cite{li2025zara}} \\
Reasoning & HAR average & Macro-F1 & \textbf{ZARA (Gemini-2.0-Flash)} \cite{li2025zara} & 0.814 \\
Reasoning & HAR average & Macro-F1 & UniMTS \cite{zhang2024unimts} & 0.321 \\
Reasoning & HAR average & Macro-F1 & IMU2CLIP \cite{moon2023imu2clip} & 0.179 \\
Reasoning & HAR average & Macro-F1 & ImageBind \cite{girdhar2023imagebind} & 0.142 \\
Reasoning & HAR average & Macro-F1 & Gemini Plot & 0.135 \\
Reasoning & HAR average & Macro-F1 & Gemini Table & 0.130 \\
Reasoning & HAR average & Macro-F1 & Gemini Text & 0.129 \\
Reasoning & HAR average & Macro-F1 & HARGPT Text \cite{ji2024hargpt} & 0.104 \\
Reasoning & HAR average & Macro-F1 & IMUGPT \cite{leng2023generating} & 0.100 \\
Reasoning & HAR average & Macro-F1 & HARGPT Plot \cite{ji2024hargpt} & 0.099 \\
Reasoning & HAR average & Macro-F1 & NormWear \cite{luo2024normwear} & 0.077 \\

\midrule
\multicolumn{5}{l}{\textit{KPI anomaly detection} \cite{ren2019time}} \\
Anomaly detection & KPI & F1 & \textbf{ARGOS (GPT-4o)} \cite{gu2501argos} & 0.897 \\
Anomaly detection & KPI & F1 & LSTMAD \cite{malhotra2015lstm} & 0.819 \\
Anomaly detection & KPI & F1 & FCVAE \cite{wang2024revisiting} & 0.818 \\
Anomaly detection & KPI & F1 & \textbf{MERIT (Llama3.1-8B)} \cite{zhou2025merit} & 0.815 \\
Anomaly detection & KPI & F1 & TimesURL \cite{liu2024timesurl} & 0.803 \\
Anomaly detection & KPI & F1 & TS2Vec \cite{yue2022ts2vec} & 0.791 \\
Anomaly detection & KPI & F1 & SR \cite{ren2019time} & 0.785 \\
Anomaly detection & KPI & F1 & DONUT \cite{xu2018donut} & 0.779 \\
Anomaly detection & KPI & F1 & DSPOT \cite{siffer2017anomaly} & 0.768 \\
Anomaly detection & KPI & F1 & SPOT \cite{siffer2017anomaly} & 0.751 \\
Anomaly detection & KPI & F1 & AutoRegression & 0.668 \\
Anomaly detection & KPI & F1 & TFAD \cite{zhang2022tfad} & 0.564 \\
Anomaly detection & KPI & F1 & AnomalyTransformer \cite{xu2021anomaly} & 0.282 \\
\bottomrule
\end{tabular}}
\end{table*}

\begin{table*}[p]
\centering
\caption{Comparable quantitative evidence across time-series tasks (Part 2). Agent methods are highlighted in bold.}
\label{tab:comparable-all-tasks-continued}
\scriptsize
\renewcommand{\arraystretch}{0.86}
\setlength{\tabcolsep}{3pt}
\resizebox{\textwidth}{!}{%
\begin{tabular}{p{0.16\linewidth} p{0.16\linewidth} p{0.11\linewidth} p{0.34\linewidth} p{0.08\linewidth}}
\toprule
\textbf{Task} & \textbf{Dataset} & \textbf{Metric} & \textbf{Method} & \textbf{Score} \\
\midrule

\multicolumn{5}{l}{\textit{Yahoo anomaly detection} \cite{laptev2015yahoo}} \\
Anomaly detection & Yahoo & F1 & \textbf{MERIT (Llama3.1-8B)} \cite{zhou2025merit} & 0.905 \\
Anomaly detection & Yahoo & F1 & TimesURL \cite{liu2024timesurl} & 0.892 \\
Anomaly detection & Yahoo & F1 & TS2Vec \cite{yue2022ts2vec} & 0.885 \\
Anomaly detection & Yahoo & F1 & SR \cite{ren2019time} & 0.879 \\
Anomaly detection & Yahoo & F1 & DONUT \cite{xu2018donut} & 0.873 \\
Anomaly detection & Yahoo & F1 & DSPOT \cite{siffer2017anomaly} & 0.861 \\
Anomaly detection & Yahoo & F1 & SPOT \cite{siffer2017anomaly} & 0.847 \\
Anomaly detection & Yahoo & F1 & \textbf{ARGOS (GPT-4o)} \cite{gu2501argos} & 0.810 \\
Anomaly detection & Yahoo & F1 & TFAD \cite{zhang2022tfad} & 0.773 \\
Anomaly detection & Yahoo & F1 & AutoRegression & 0.526 \\
Anomaly detection & Yahoo & F1 & FCVAE \cite{wang2024revisiting} & 0.464 \\
Anomaly detection & Yahoo & F1 & LSTMAD \cite{malhotra2015lstm} & 0.350 \\
Anomaly detection & Yahoo & F1 & LLMAD (GPT-4-32k) \cite{liu2024llmad} & 0.142 \\

\midrule
\multicolumn{5}{l}{\textit{AAPL single-asset trading} \cite{yu2024fincon}} \\
Trading & AAPL & SR & \textbf{FinCon (GPT-4-Turbo)} \cite{yu2024fincon} & 1.597 \\
Trading & AAPL & SR & FinGPT (GPT-4-Turbo) \cite{yang2023fingpt} & 1.161 \\
Trading & AAPL & SR & B\&H & 1.107 \\
Trading & AAPL & SR & DQN & 1.048 \\
Trading & AAPL & SR & \textbf{FinAgent (GPT-4-Turbo)} \cite{zhang2024multimodal} & 1.041 \\
Trading & AAPL & SR & \textbf{FinMem (GPT-4-Turbo)} \cite{yu2025finmem} & 0.994 \\
Trading & AAPL & SR & PPO & 0.704 \\
Trading & AAPL & SR & A2C & 0.683 \\
Trading & AAPL & SR & GA (GPT-4-Turbo) \cite{park2023generative} & 0.372 \\

\midrule
\multicolumn{5}{l}{\textit{AMZN single-asset trading} \cite{yu2024fincon}} \\
Trading & AMZN & SR & \textbf{FinCon (GPT-4-Turbo)} \cite{yu2024fincon} & 0.904 \\
Trading & AMZN & SR & DQN & 0.398 \\
Trading & AMZN & SR & PPO & 0.138 \\
Trading & AMZN & SR & B\&H & 0.072 \\
Trading & AMZN & SR & GA (GPT-4-Turbo) \cite{park2023generative} & -0.199 \\
Trading & AMZN & SR & A2C & -0.444 \\
Trading & AMZN & SR & \textbf{FinMem (GPT-4-Turbo)} \cite{yu2025finmem} & -0.773 \\
Trading & AMZN & SR & \textbf{FinAgent (GPT-4-Turbo)} \cite{zhang2024multimodal} & -1.493 \\
Trading & AMZN & SR & FinGPT (GPT-4-Turbo) \cite{yang2023fingpt} & -1.810 \\

\midrule
\multicolumn{5}{l}{\textit{Jinan 1 traffic signal control} \cite{lai2025llmlight}} \\
Traffic control & Jinan 1 & ATT & \textbf{LightGPT (Llama2-13B)} \cite{lai2025llmlight} & 274.03 \\
Traffic control & Jinan 1 & ATT & Advanced-CoLight \cite{zhang2022expression} & 274.67 \\
Traffic control & Jinan 1 & ATT & \textbf{LightGPT (Llama3-8B)} \cite{lai2025llmlight} & 275.10 \\
Traffic control & Jinan 1 & ATT & GPT-4 & 275.26 \\
Traffic control & Jinan 1 & ATT & Efficient-CoLight \cite{zhang2022expression} & 277.11 \\
Traffic control & Jinan 1 & ATT & CoLight \cite{wei2019colight} & 279.60 \\
Traffic control & Jinan 1 & ATT & Maxpressure & 281.58 \\
Traffic control & Jinan 1 & ATT & FixedTime & 481.79 \\
Traffic control & Jinan 1 & ATT & Random & 597.62 \\

\midrule
\multicolumn{5}{l}{\textit{Open-TI Config1 traffic signal control} \cite{da2024open}} \\
Traffic control & Open-TI Config1 & ATT & \textbf{Open-TI (GPT-4.0)} \cite{da2024open} & 103.46 \\
Traffic control & Open-TI Config1 & ATT & PressLight \cite{wei2019presslight} & 107.39 \\
Traffic control & Open-TI Config1 & ATT & DQN & 162.19 \\
Traffic control & Open-TI Config1 & ATT & SOTL & 218.06 \\
Traffic control & Open-TI Config1 & ATT & FixedTime & 552.72 \\
\bottomrule
\end{tabular}}
\end{table*}
\fi

%% file: appendix_02_methods.tex
\ifshowappendixfloats\else
\section{Method Summary}
\label{sec:appendix_table1}

Table~\ref{tab:taxonomy_overview} summarizes the representative methods covered in this survey under our problem-driven taxonomy. For each method, we report its publication year and venue or source, together with three agent-oriented dimensions: architecture, tools, and memory. Detailed discussions are provided in Section~\ref{sec:taxonomy}.
\fi

\ifshowappendixfloats
\begin{table*}[p]
  \caption{Summary of representative LLM-based agentic methods for time-series tasks. Each method is grouped by \textbf{Problem Type} and further characterized by its publication year, venue or source, architecture, tools, and memory.}
  \label{tab:taxonomy_overview}
  \centering
  \tiny
  \setlength{\tabcolsep}{2pt}
  \renewcommand{\arraystretch}{0.85}
  \resizebox{\textwidth}{!}{
  \begin{tabular}{>{\raggedright\arraybackslash}p{2.2cm}>{\raggedright\arraybackslash}p{0.7cm}>{\raggedright\arraybackslash}p{0.9cm}>{\raggedright\arraybackslash}p{1.6cm}>{\raggedright\arraybackslash}p{2.4cm}>{\raggedright\arraybackslash}p{3.6cm}>{\raggedright\arraybackslash}p{1.8cm}}
    \toprule
    \textbf{Method} & \textbf{Year} & \textbf{Source} & \textbf{Problem Type} & \textbf{Architecture} & \textbf{Tools} & \textbf{Memory} \\
    \midrule
    TimeSeriesScientist \cite{zhao2025timeseriesscientist} & 2025 & arXiv & Forecasting & Multi (Cooperative) & Stat./ML Models & Evidence Logs \\
    TimeXL \cite{jiang2025timexl} & 2025 & NeurIPS & Forecasting & Multi (Cooperative) & None & Evidence Logs, Pattern Library \\
    TimeCAP \cite{lee2025timecap} & 2025 & AAAI & Forecasting & Multi (Cooperative) & Data Processing Tools & Pattern Library \\
    NewsTSForecasting \cite{zhang2025can} & 2025 & arXiv & Forecasting & Multi (Competitive) & Search \& Retrieval APIs & Evidence Logs, Analysis Strategies \\
    TRACE \cite{chen2025trace} & 2025 & IEEE TAI & Forecasting & Multi (Competitive) & None & Evidence Logs \\
    FLAIRR-TS \cite{jalori2025flairr} & 2025 & EMNLP Findings & Forecasting & Multi (Cooperative) & None & Evidence Logs, Analysis Strategies \\
    Nexus \cite{das2026nexus} & 2026 & arXiv & Forecasting & Multi (Cooperative) & Data Processing Tools & Evidence Logs \\
    CastFlow \cite{pan2026castflow} & 2026 & arXiv & Forecasting & Multi (Mixed) & Data Processing Tools, Stat./ML Models & Evidence Logs, Pattern Library \\
    TS-Agent \cite{liu2025ts} & 2025 & NeurIPS Wkshp. & Reasoning & Single & Data Processing Tools, Stat./ML Models & Evidence Logs \\
    Agentic-RAG \cite{ravuru2024agentic} & 2024 & KDD UC & Reasoning & Multi (Cooperative) & None & Pattern Library \\
    TS-Reasoner \cite{ye2024domain} & 2024 & arXiv & Reasoning & Single & Database APIs, Data Processing Tools, Stat./ML Models & Evidence Logs \\
    ZARA \cite{li2025zara} & 2026 & ACL & Reasoning & Multi (Cooperative) & Database APIs, Data Processing Tools, Stat./ML Models & Pattern Library, Analysis Strategies \\
    CLIMATEAGENT \cite{kim2025climateagent} & 2025 & arXiv & Reasoning & Multi (Cooperative) & Database APIs, Data Processing Tools, Simulators, Solvers \& Optimizers & Evidence Logs \\
    FETA \cite{sui2025feta} & 2025 & arXiv & Reasoning & Multi (Competitive) & Data Processing Tools, Search \& Retrieval APIs & Pattern Library \\
    \midrule
    TESSA \cite{lin2024decoding} & 2026 & EACL & Augmentation & Multi (Cooperative) & Data Processing Tools & None \\
    DCATS \cite{yeh2025empowering} & 2025 & arXiv & Augmentation & Single & Stat./ML Models & Evidence Logs \\
    MERIT \cite{zhou2025merit} & 2025 & ACL & Augmentation & Multi (Cooperative) & Data Processing Tools, Stat./ML Models & None \\
    ChatTS \cite{xie2024chatts} & 2025 & PVLDB & Synthesis & Multi (Cooperative) & None & None \\
    GenAI4RiskModeling \cite{joshi2025using} & 2025 & SSRN & Synthesis & Multi (Cooperative) & Database APIs, Data Processing Tools, Stat./ML Models & None \\
    BRIDGE \cite{li2025bridge} & 2025 & ICML & Synthesis & Multi (Mixed) & Search \& Retrieval APIs, Stat./ML Models & Pattern Library \\
    \midrule
    AD-AGENT \cite{yang2025ad} & 2025 & AACL & Detection & Multi (Cooperative) & Data Processing Tools, Stat./ML Models & Evidence Logs, Analysis Strategies \\
    ARGOS \cite{gu2501argos} & 2025 & arXiv & Detection & Multi (Mixed) & Data Processing Tools, Stat./ML Models & Evidence Logs \\
    SLEP \cite{wang2025large} & 2026 & ESWA & Detection & Single & Data Processing Tools & Evidence Logs, Analysis Strategies \\
    CALM \cite{devireddy2025calm} & 2025 & arXiv & Detection & Single & Data Processing Tools, Stat./ML Models & None \\
    SAGE \cite{kang2026sage} & 2026 & arXiv & Detection & Multi (Cooperative) & Data Processing Tools, Stat./ML Models & Evidence Logs \\
    LEMAD \cite{ji2025lemad} & 2025 & Electron. & Diagnosis & Multi (Cooperative) & Data Processing Tools, Stat./ML Models & Evidence Logs \\
    AgentFM \cite{zhang2025agentfm} & 2025 & FSE & Diagnosis & Multi (Cooperative) & Database APIs, Data Processing Tools, Stat./ML Models & None \\
    LLM-TSFD \cite{zhang2025llm} & 2025 & ESWA & Diagnosis & Single & Database APIs, Data Processing Tools, Search \& Retrieval APIs, Stat./ML Models & Pattern Library \\
    \midrule
    ElliottAgents \cite{chudziak2024elliottagents,wawer2025integrating} & 2024 & PACLIC & Trading & Multi (Cooperative) & Data Processing Tools, Stat./ML Models & Pattern Library \\
    TradingAgents \cite{xiao2024tradingagents} & 2024 & arXiv & Trading & Multi (Mixed) & Data Processing Tools, Search \& Retrieval APIs & Evidence Logs \\
    FinCon \cite{yu2024fincon} & 2024 & NeurIPS & Trading & Multi (Cooperative) & Database APIs, Data Processing Tools, Search \& Retrieval APIs & Evidence Logs, Pattern Library, Analysis Strategies \\
    FinAgent \cite{zhang2024multimodal} & 2024 & KDD & Trading & Multi (Cooperative) & Data Processing Tools & Pattern Library, Analysis Strategies \\
    FactorMAD \cite{duan2025factormad} & 2025 & ICAIF & Trading & Multi (Competitive) & Data Processing Tools, Stat./ML Models & Pattern Library \\
    FinMem \cite{yu2025finmem} & 2025 & IEEE TBD & Trading & Single & Database APIs, Data Processing Tools & Evidence Logs, Pattern Library \\
    FinArena \cite{xu2025finarena} & 2025 & arXiv & Trading & Multi (Cooperative) & Data Processing Tools, Search \& Retrieval APIs & None \\
    ATLAS \cite{papadakis2025atlas} & 2025 & arXiv & Trading & Multi (Cooperative) & Search \& Retrieval APIs, Simulators, Solvers \& Optimizers & Analysis Strategies \\
    FinRL-DeepSeek \cite{benhenda2025finrl} & 2025 & arXiv & Trading & Single & Stat./ML Models & None \\
    Open-TI \cite{da2024open} & 2024 & IJMLC & Traffic Control & Multi (Cooperative) & Database APIs, Data Processing Tools, Stat./ML Models & None \\
    LLMTraveler \cite{wang2025agentic} & 2025 & TR-C & Traffic Control & Single & None & Evidence Logs \\
    LLMLight \cite{lai2025llmlight} & 2025 & KDD & Traffic Control & Single & Simulators, Solvers \& Optimizers & None \\
    CoLLMLight \cite{yuan2025collmlight} & 2026 & ICLR & Traffic Control & Multi (Mixed) & Data Processing Tools, Simulators, Solvers \& Optimizers & Evidence Logs \\
    HeraldLight \cite{guo2025heraldlight} & 2025 & arXiv & Traffic Control & Multi (Mixed) & Stat./ML Models, Simulators, Solvers \& Optimizers & Evidence Logs \\
    Traffic-R1 \cite{zou2025trafficr1} & 2025 & arXiv & Traffic Control & Single & Simulators, Solvers \& Optimizers & Analysis Strategies \\
    Virtual Traffic Police \cite{wei2026virtualtraffic} & 2026 & arXiv & Traffic Control & Single & Search \& Retrieval APIs, Simulators, Solvers \& Optimizers & Evidence Logs \\
    CuraLight \cite{guo2026curalight} & 2026 & arXiv & Traffic Control & Multi (Competitive) & Simulators, Solvers \& Optimizers & Evidence Logs \\
    Grid-Agent \cite{zhang2025grid} & 2025 & arXiv & Grid Control & Multi (Cooperative) & Simulators, Solvers \& Optimizers & Evidence Logs \\
    GridMind \cite{jin2025gridmind} & 2025 & SC Workshops & Grid Control & Multi (Cooperative) & Simulators, Solvers \& Optimizers & Evidence Logs \\
\bottomrule
  \end{tabular}}
\end{table*}

\begin{table*}[p]
  \caption{Within-family counts and percentages of architecture, tool, and memory choices among the 47 systems summarized in Table~\ref{tab:taxonomy_overview}. Tool and memory categories are non-exclusive.}
  \label{tab:design_choice_distribution}
  \centering
  \scriptsize
  \setlength{\tabcolsep}{3pt}
  \renewcommand{\arraystretch}{1.05}
  \resizebox{\textwidth}{!}{
  \begin{tabular}{>{\raggedright\arraybackslash}p{2.8cm}>{\raggedright\arraybackslash}p{4.4cm}>{\raggedright\arraybackslash}p{5.7cm}>{\raggedright\arraybackslash}p{4.4cm}}
    \toprule
    \textbf{Task Family} & \textbf{Architecture} & \textbf{Tools} & \textbf{Memory} \\
    \midrule
    Forecasting \& Reasoning ($n=14$) &
    Single: 2 (14.3\%); Multi (Cooperative): 8 (57.1\%); Multi (Competitive): 3 (21.4\%); Multi (Mixed): 1 (7.1\%) &
    Database APIs: 3 (21.4\%); Search \& Retrieval APIs: 2 (14.3\%); Data Processing Tools: 8 (57.1\%); Stat./ML Models: 5 (35.7\%); Simulators, Solvers \& Optimizers: 1 (7.1\%); None: 4 (28.6\%) &
    Evidence Logs: 10 (71.4\%); Pattern Library: 6 (42.9\%); Analysis Strategies: 3 (21.4\%); None: 0 (0\%) \\
    \midrule
    Augmentation \& Synthesis ($n=6$) &
    Single: 1 (16.7\%); Multi (Cooperative): 4 (66.7\%); Multi (Competitive): 0 (0\%); Multi (Mixed): 1 (16.7\%) &
    Database APIs: 1 (16.7\%); Search \& Retrieval APIs: 1 (16.7\%); Data Processing Tools: 3 (50.0\%); Stat./ML Models: 4 (66.7\%); Simulators, Solvers \& Optimizers: 0 (0\%); None: 1 (16.7\%) &
    Evidence Logs: 1 (16.7\%); Pattern Library: 1 (16.7\%); Analysis Strategies: 0 (0\%); None: 4 (66.7\%) \\
    \midrule
    Anomaly Detection \& Diagnosis ($n=8$) &
    Single: 3 (37.5\%); Multi (Cooperative): 4 (50.0\%); Multi (Competitive): 0 (0\%); Multi (Mixed): 1 (12.5\%) &
    Database APIs: 2 (25.0\%); Search \& Retrieval APIs: 1 (12.5\%); Data Processing Tools: 8 (100\%); Stat./ML Models: 7 (87.5\%); Simulators, Solvers \& Optimizers: 0 (0\%); None: 0 (0\%) &
    Evidence Logs: 5 (62.5\%); Pattern Library: 1 (12.5\%); Analysis Strategies: 2 (25.0\%); None: 2 (25.0\%) \\
    \midrule
    Decision Support ($n=19$) &
    Single: 6 (31.6\%); Multi (Cooperative): 8 (42.1\%); Multi (Competitive): 2 (10.5\%); Multi (Mixed): 3 (15.8\%) &
    Database APIs: 3 (15.8\%); Search \& Retrieval APIs: 5 (26.3\%); Data Processing Tools: 9 (47.4\%); Stat./ML Models: 5 (26.3\%); Simulators, Solvers \& Optimizers: 9 (47.4\%); None: 1 (5.3\%) &
    Evidence Logs: 10 (52.6\%); Pattern Library: 5 (26.3\%); Analysis Strategies: 4 (21.1\%); None: 4 (21.1\%) \\
    \bottomrule
  \end{tabular}}
\end{table*}
\fi